\documentclass{article} 
\usepackage{iclr2026_conference,times}

\usepackage{amsmath,amsfonts,bm}

\def\eqref#1{equation~\ref{#1}}

\def\1{\bm{1}}

\DeclareMathAlphabet{\mathsfit}{\encodingdefault}{\sfdefault}{m}{sl}
\SetMathAlphabet{\mathsfit}{bold}{\encodingdefault}{\sfdefault}{bx}{n}

\DeclareMathOperator*{\minimize}{Minimize}

\newcommand{\cu}{{\textcolor{blue}{u}}}

\newcommand{\cy}{{\textcolor{brown}{y}}}

\usepackage{hyperref}
\usepackage{url}
\usepackage{subfigure}
\usepackage{subcaption}
\usepackage{booktabs} 
\usepackage{mdframed}
\usepackage{xcolor}
\usepackage{paralist}
\usepackage{enumitem}
\usepackage{wrapfig}
\usepackage{adjustbox}
\usepackage{algpseudocode}
\usepackage{algcompatible}
\usepackage{algorithm}
\usepackage{multirow}
\usepackage{multicol}
\usepackage{array, makecell}
\usepackage{mathtools}

\usepackage{float}
\newfloat{model}{thp}{lop}
\floatname{model}{Model}

\usepackage{multirow,mdframed}
\mdfsetup{skipabove=0pt,skipbelow=0pt}
\mdfdefinestyle{model}{%
  innertopmargin=0pt,
  innerbottommargin=0pt,
  innerleftmargin=0pt,
  innerrightmargin=0pt,
  skipbelow=0pt,%
  skipabove=0pt,
  splitbottomskip=0pt,
  splittopskip=0pt,
  leftmargin =0pt,%
    rightmargin=0pt,
  splittopskip=0pt,
    usetwoside=false,
}

\title{Learning to Solve \\ Optimization Problems Constrained \\ with Partial Differential Equations}

\author{Yusuf Güven
\thanks{Equal Contribution
\\
\\
\! \! The implementation of the methods used to reproduce the experiments described in this work con be found at \url{https://github.com/vdvf96/PDE-OP.}  } \\
Istanbul Technical University\\
Control and Automation Engineering Department\\
Istanbul,  Turkey \\
\texttt{ngu6fh@virginia.edu} \\
\And
Vincenzo Di Vito$^*$ and Ferdinando Fioretto \\
University of Virginia \\
Department of Computer Science \\
Charlottesville, USA \\
\texttt{\{eda8pc,fioretto\}@virginia.edu} \\
}

\iclrfinalcopy 
\begin{document}

\maketitle

\begin{abstract}

Partial differential equation (PDE)-constrained optimization arises in many scientific and engineering domains, such as energy systems, fluid dynamics, and material design. In these problems, the decision variables (e.g., control inputs or design parameters) are tightly coupled with the PDE state variables, and the feasible set is implicitly defined by the governing PDE constraints. This coupling makes the problems computationally demanding, as it requires handling high-dimensional discretizations and dynamic constraints.
To address these challenges, this paper introduces a learning-based framework that integrates a dynamic predictor with an optimization surrogate. The dynamic predictor, a novel time-discrete Neural Operator~\citep{Lu_2021}, efficiently approximates system trajectories governed by PDE dynamics, while the optimization surrogate leverages proxy optimizer techniques~\citep{kotary2021end} to approximate the associated optimal decisions. This dual-network design enables real-time approximation of optimal strategies while explicitly capturing the coupling between decisions and PDE dynamics.
We validate the proposed approach on benchmark PDE-constrained optimization tasks including Burgers’ equation, heat equation, and voltage regulation, and demonstrate that it achieves solution quality comparable to classical control-based algorithms such as the Direct Method and Model Predictive Control (MPC), while providing up to four orders of magnitude improvement in computational speed. 
\end{abstract}



\author{Antiquus S.~Hippocampus, Natalia Cerebro \& Amelie P. Amygdale \thanks{$^*$Equal Contribution} \\
Department of Computer Science\\
Cranberry-Lemon University\\
Pittsburgh, PA 15213, USA \\
\texttt{\{hippo,brain,jen\}@cs.cranberry-lemon.edu} \\
\And
Ji Q. Ren \& Yevgeny LeNet \\
Department of Computational Neuroscience \\
University of the Witwatersrand \\
Joburg, South Africa \\
\texttt{\{robot,net\}@wits.ac.za} \\
\AND
Coauthor \\
Affiliation \\
Address \\
\texttt{email}
}

%

\iclrfinalcopy 
\raggedbottom
\sloppy

\maketitle

\section{Introduction}

Partial differential equations (PDEs) are central to modeling a wide range of physical and engineering systems, capturing phenomena that evolve jointly over space and time. Applications include fluid transport, which can be modeled with Burgers’ equations which are key in fluid mechanics and gas dynamics, thermal processes captured by heat equations, often used in financial mathematics and image analysis, and voltage dynamics in power systems. Optimizing decisions in such domains requires solving \emph{PDE-constrained optimization problems}, where control or design objectives must be optimized while ensuring consistency with the underlying governing PDE dynamics.

Despite their importance, PDE-constrained optimization remains computationally intractable. Traditional approaches typically rely on finite element and finite difference discretizations combined with adjoint-based optimization. These methods often suffer from high dimensionality, high computational cost, and convergence difficulties when applied to nonlinear dynamics. For example, when applied to the nonlinear Burgers’ equations, our experiments show that the runtimes of the adjoint-based and Nonlinear MPC approaches run in approximately $120 \;s$ and $878 \; s$; such computational cost hinder their applicability in large-scale or real-time decision-making tasks.


In various classes of continuous optimization tasks not constrained by PDEs, proxy optimizers \citep{kotary2021end} have recently emerged as a powerful tool to quickly find approximate solutions. 
In parallel, researchers have focused on developing a variety of surrogates \citep{raissi2017physics, kovachki2024neuraloperatorlearningmaps, li2021fourier, chen2019neural} for capturing system dynamics. Our approach is a unique combination of elements from both these areas, and consists of a dual-network architecture  
to tackle PDE-constrained optimization problems where both the state dynamics and the control decisions evolve in space and time. 

\textbf{Contributions.} 
Specifically, the paper makes the following contributions:  
{\bf (1)} It introduces a novel learning-based method to efficiently approximate solutions of PDE-constrained optimization problems. Our approach, PDE-OP (Optimization Proxy), consists of a dual-network architecture in which a dynamic predictor captures the system dynamics, while a surrogate controller approximates the optimal control actions. These components are trained in a synergistic manner in a self-supervised fashion using a primal–dual method.  
{\bf (2)} It shows that the proposed time-discrete Neural Operator accurately captures the dynamics governed by PDEs.
{\bf (3)} It demonstrates that the proposed framework achieves decision quality comparable to state-of-the-art numerical solvers, such as Model Predictive Control and the Adjoint-sensitivity method, while providing several orders of magnitude improvements in computational speed. The approach is validated on a set of widely adopted PDE-constrained optimization tasks,  including Burgers’, heat, and voltage equations.
\textit{We believe that these results could boost the adoption of learning-based methods for control problems that require high-fidelity dynamic modeling and near real-time optimal decision-making.}



\section{Related Work}
\label{app:related_work}
A central paradigm for decision-making coupled with dynamical systems is \emph{Model Predictive Control (MPC)} \citep{mayne2000constrained}. It repeatedly solves an optimization problem over a finite time horizon using predicted system states. While MPC provides high-quality control decisions, its reliance on repeatedly solving an optimization problem makes it computationally intractable and limits its applicability in real-time settings. In a similar fashion, exact PDE-constrained optimization methods rely on time–space discretization of the governing equations, after which the resulting high-dimensional system is incorporated into an optimization problem \citep{hinze2008optimization, troeltzsch2010optimal}. While theoretically well-founded, these approaches become quickly impractical as the problem size grows, especially in the presence of nonlinear dynamics or when real-time feasibility is required. To mitigate these limitations, surrogate models have been developed for reduced-order adjoint solutions \citep{zahr2014progressive}, sensitivity PDEs \citep{dihlmann2015certified}, and learning PDE operators \citep{Hwang_2022}, offering significant computational advantages. Nevertheless, these methods often face trade-offs between accuracy and efficiency and limited adaptability across operating regimes\citep{kovachki2024neuraloperatorlearningmaps, Lu_2021, willcox2021physics}.

In parallel, to approximate solutions of constrained optimization problems in real-time setting, machine learning researchers have developed a class of methods known as \emph{proxy optimizers}. These methods employ machine learning models, to learn mappings from problem parameters or system configurations to optimal decisions, thereby significantly accelerating inference \citep{kotary2021end} compared to classical optimization solvers. Both supervised approaches \citep{fioretto2020lagrangian}, which rely on precomputed optimal solutions, and self-supervised ones \citep{park2023self}, which exploit knowledge of the problem structure, have been proposed, showing promising results. A recurring challenge in this setting is ensuring feasibility: learned solutions may violate problem constraints. This issue has been tackled through penalty-based training \citep{fioretto2020lagrangian}, implicit-layer formulations that embed constraints directly into the model architecture \citep{donti2020dc3}, or post-processing projections to restore feasibility \citep{kotary2024learningjointmodelsprediction}.

A large body of recent work has also focused on \emph{learning surrogates for PDE dynamics}. Physics-Informed Neural Networks  \citep{2019JCoPh.378..686R} directly encode PDE residuals into the loss function, enabling solutions that adhere to the governing equations.
Neural operator approaches, such as the Fourier Neural Operator \citep{li2021fourier} and DeepONet \citep{Lu_2021}, learn mappings between infinite-dimensional function spaces, which allows them to approximate solution operators of \textit{families of PDEs}, with strong generalization across different boundary conditions and coefficients \citep{kovachki2024neuraloperatorlearningmaps}. Other operator-learning frameworks \citep{li2020multipole, bhattacharya2021model} and kernel-based methods have also been developed. While these approaches excel at reproducing PDE solutions, they are typically not designed to incorporate decision variables or to directly solve PDE-constrained optimization problems.
The closest related work is \citep{divito2024learningsolvedifferentialequation}, where the authors propose a learning-based surrogate solver to tackle ODE or SDE-constrained optimization tasks. 
\textit{Our approach goes further, by integrating operator-learning surrogates within the optimization pipeline, thereby coupling the representation of PDE dynamics with proxy optimizers and enabling real-time solutions to PDE-constrained optimization tasks.}

\section{Settings and Goals}
\label{sec:settings}
This paper focuses on optimization problem constrained by a system of partial 
differential equations:
\begin{subequations}
\label{eq:1}
\begin{align}
\minimize_{\,\cu:Q\to\mathbb{R}} \;\; 
& \overbrace{L\!\big(\cy(T,\cdot)\big) \;+\; \int_0^T\!\!\int_{\Omega}
\Phi\!\big(\cu(t,{x}), \cy(t,{x})\big)\, d{x}\, dt + \int_0^T\!\!\int_{\Omega}
\ \| u (t,x) \| ^2_2 d{x}\, dt
}^{J(\cu(t, {x}), \cy(t, {x}))}
\label{eq:1a}\\[2pt]
\text{s.t.}\quad 
& \partial_t \cy(t,{x}) \;=\; {\mathcal{I}}_{t}\!\big(\cy(t,{x}), \cu(t,{x}), t, {x}\big),
\; \; \; \; \; (t,{x}) \in Q, \label{eq:1b}\\
& \partial_{{x}} \cy(t,{x}) \;=\; {\mathcal{I}}_{x}\!\big(\cy(t,{x}), \cu(t,{x}), t, {x}\big),
\; \; \; \; (t,{x}) \in Q, \label{eq:1c}\\
& \cy(t_0,{x}) \;=\; \cy_{t_0}({x}),
\quad {x} \in \Omega, \label{eq:1d}\\
& \cy(t,{x}) \;=\; \cy_{\partial}(t,{x}),
\quad \quad  (t,{x}) \in [t_0,T]\times \partial\Omega, \label{eq:1e}\\
& \bm{g}(\cu(t,x),\cy(t,x) \;\le\; \bm{0}, \quad 
  \bm{h}(\cu(t,x),\cy(t,x)) \;=\; \bm{0},
\quad (t,{x}) \in Q. \label{eq:1f}
\end{align}
\end{subequations}
\noindent
In this formulation, the control action is denoted by \(\cu:Q\to\mathbb{R}\), where $Q=[t_0,T]\times\Omega$ is the space–time domain, with $t_0, T \in \mathbb{R}$ being the initial time instant and the time horizon, respectively, and $\Omega$ being the spatial domain, e.g., if $x \in \mathbb{R}$, then $\Omega \subseteq \mathbb{R}$. The system state is represented by \(\cy:Q\to\mathbb{R}\). $t$ and ${x}$ denote the independent temporal and spatial variable, respectively. The optimization objective~(\ref{eq:1a}) combines a terminal cost \(L\big(\cy(T,\cdot)\big)\), which depends on the state at the final time $T$, with a running cost $\Phi$, integrated over time and space, and an effort cost, which is the energy used to control the system.  

The system dynamics are governed by two 
PDE constraints. Equation~(\ref{eq:1b}) models the temporal evolution of the state, while~(\ref{eq:1c}) captures spatial dependencies, for example through diffusion–reaction relations. Initial conditions~(\ref{eq:1d}) prescribe the state distribution at $t=t_0$, while boundary conditions~(\ref{eq:1e}) specify the state behavior on the spatial boundary $\partial \Omega$. Finally,~(\ref{eq:1f}) represents additional inequality and equality constraints on \((\cu, \cy)\), which may encode feasibility requirements or application-specific restrictions.

\begin{wrapfigure}[13]{r}{0.5\textwidth}  
\vspace{-26pt}
    \centering
    \includegraphics[width=0.85\linewidth]{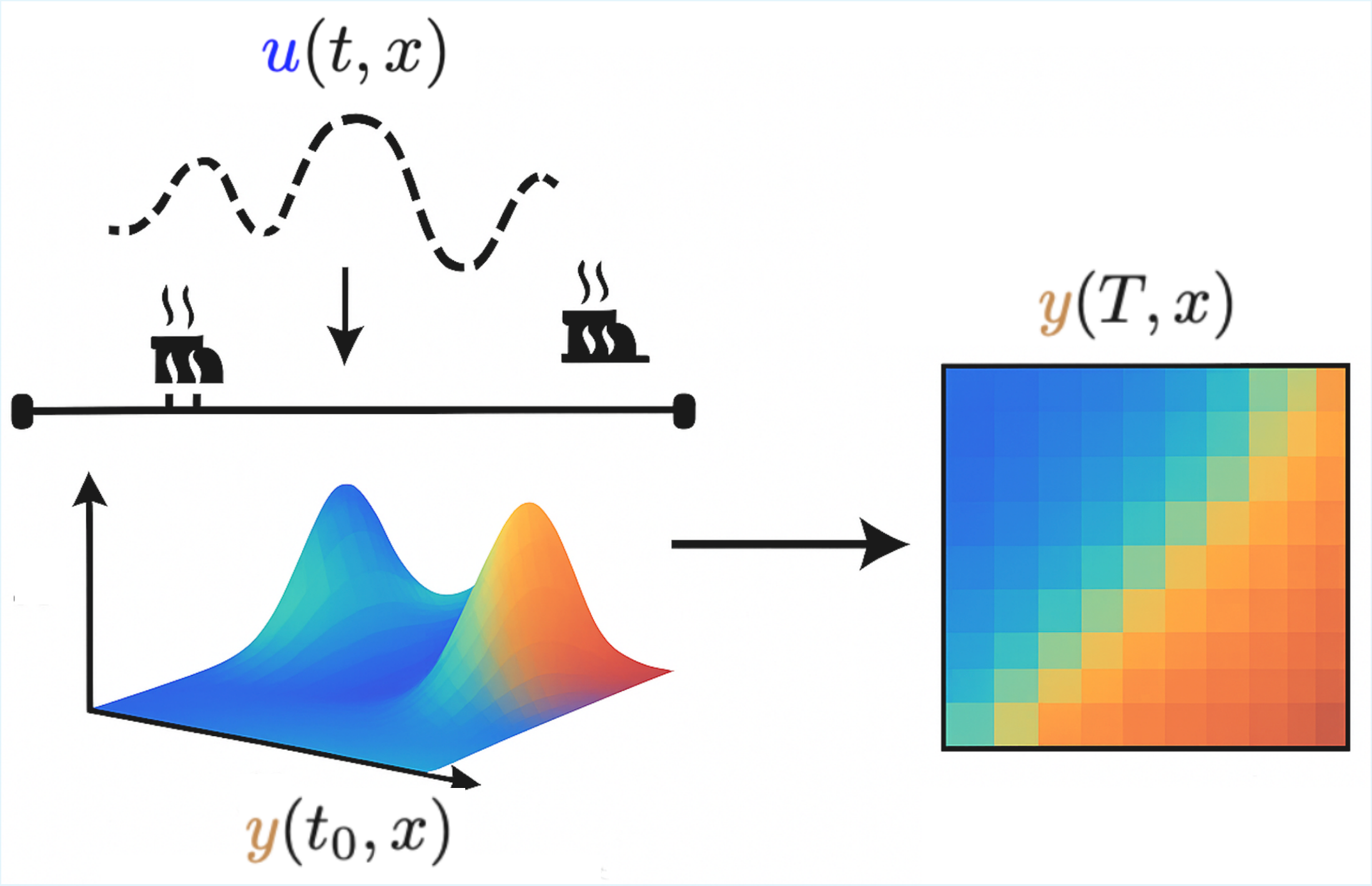}
    \vspace{-4pt}
    \caption{Temperature regulation process: control actions \(\cu(t,x)\) steer the system from the initial state \(\cy(t_0,x)\) towards a target state \(\cy(T,x)\).}
    \label{fig:power_sys_example}
\vspace{-10pt}
\end{wrapfigure}

\noindent\textbf{Temperature control example.}
In the context of thermal regulation problems, the decision variable \(\cu(t, x)\) represents localized actuators applied along the medium, while the state variable $\cy(t, x)$ describes the temperature distribution over space and time. These two components are tightly coupled: the control inputs directly enter the governing PDE as forcing terms, shaping the spatio-temporal evolution of the state, while the optimization objective is evaluated on the resulting state trajectory. System dynamics~(\ref{eq:1b}–\ref{eq:1c}) follow a controlled reaction–diffusion equation, which models heat diffusion, dissipation, and externally applied control. The initial condition~(\ref{eq:1d}) specifies the temperature distribution at $t=t_0$, and the boundary condition~(\ref{eq:1e}) enforces zero-flux (Neumann) boundaries, preventing heat transfer across the domain boundary. The objective~\ref{eq:1a} combines a terminal cost to match the target temperature profile, a running cost for deviations along the horizon, and a control effort penalty. A simplified illustration is provided in Fig.\ref{fig:power_sys_example}, while a full mathematical description is given in Section~\ref{sec:Heat}.


\textbf{Challenges.} While fundamental for many applications, 
Problem (\ref{eq:1}) presents three key 
challenges:
\begin{enumerate}
[itemsep=0pt,leftmargin=*,topsep=-4pt,parsep=0pt]
\item Finding optimal solutions to Problem (\ref{eq:1}) is computationally intractable. Even without the differential equation constraints, the decision version of 
the problem alone is NP-hard in general.
\item Achieving high-quality approximations of the system dynamics (\ref{eq:1b})-(\ref{eq:1c}) in near real-time, poses the second significant challenge. The non-linearity of these dynamics further complicate the task.

\item Finally, the integration of the system dynamics 
in the optimization framework renders traditional numerical methods impractical for real-time applications. 
\end{enumerate}  

\section{PDE-Optimization Proxy}

\begin{figure}[t!]
\centering
\includegraphics[width=0.9\linewidth]{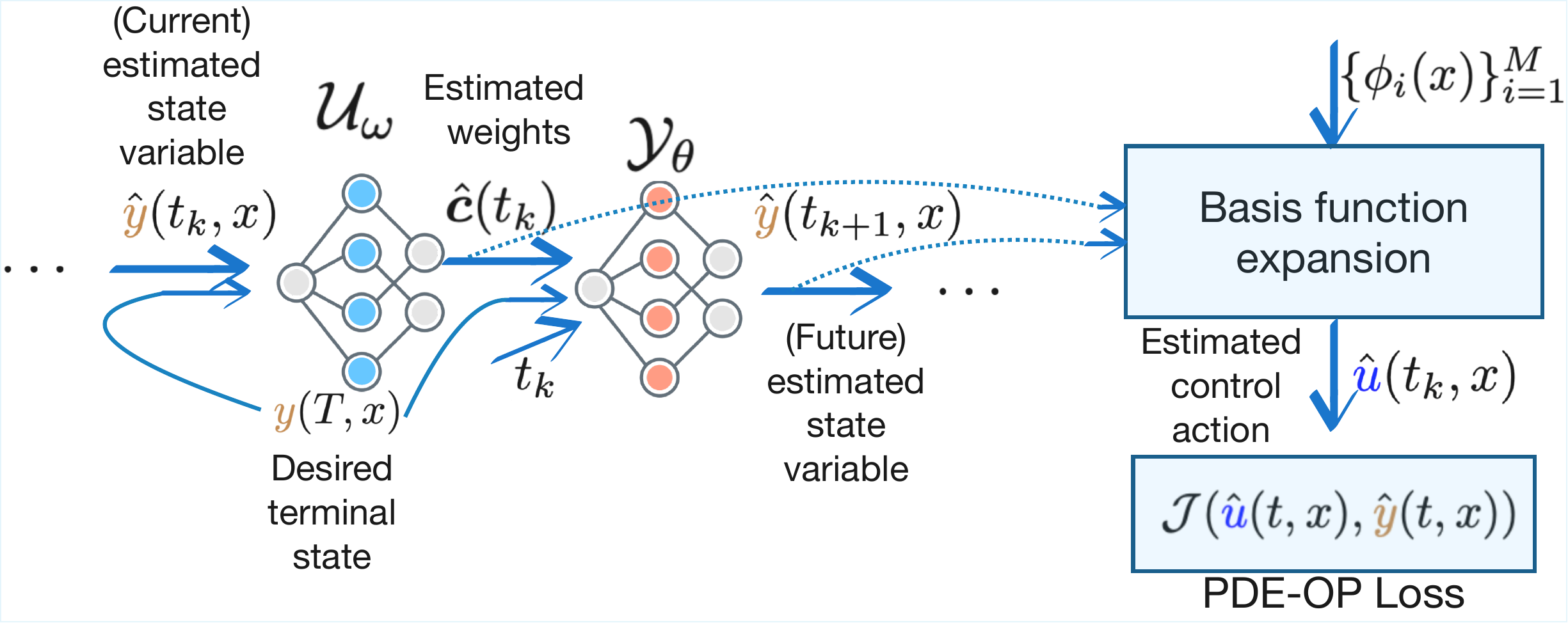}
\caption{At each time instant $t_k$, PDE-OP uses a dual network architecture consisting of a proxy optimization model $\mathcal{U}_\omega$ to estimate the basis functions weights  $\bm{{c}}(t_k)$ and construct the control action $\hat{{\cu}}(t_k,x)$, and a neural-DE model $\mathcal{Y}_\theta$ to estimate the state-variables $\hat{{\cy}}(t_{k+1},x)$, with the objective function $\mathcal{J}(\hat{{\cu}}(t,x), \hat{{\cy}}(t,x))$ capturing the overall loss.}
\label{fig:DE-OP_scheme}
\end{figure}

To tackle the challenges discussed above, we introduce \emph{PDE-Optimization Proxy} (PDE-OP): a fully differentiable surrogate for PDE-constrained optimization. 
PDE-OP adopts a dual-network architecture: the first network, $\mathcal{U}_\omega$, approximates the optimal decision variable 
${u}^*(t,{x})$, 
while the second, $\mathcal{Y}_\theta$, predicts the corresponding state variables ${y}(t,{x})$ 
using a 
discrete-time neural operator architecture. 
Here, $\omega$ and $\theta$ denote the trainable parameters of $\mathcal{U}$ and $\mathcal{Y}$, respectively. 
These two components are jointly trained in an end-to-end differentiable manner, and a schematic illustration of their interaction is provided in Figure~\ref{fig:DE-OP_scheme}. 
In what follows, we first describe each network individually and then detail their integration within the proposed learning framework.

\textbf{Learning Goal.}
PDE-OP trains on a distribution $\Pi$ of problem instances generated by different states $y(t,x)$ for various $t$ and $x$ values. For instance, in the thermal regulation example, this distribution corresponds to the variety of temperature profiles that arise from different control inputs. 
With reference to (\ref{eq:1a}), the learning objective is:
\begin{subequations}
\label{eq:goal}
\begin{align}
    \minimize_{\omega, \theta} \mathbb{E}_{{y(t,x)} \sim \Pi} 
        \left[ {\cal J}\left(\mathcal{U}_\omega(t,x,y), \mathcal{Y}_\theta( t, {x}, \hat u)
        \right) \right]
        \label{eq:2a}\\
\texttt{s.t.} \;\ \text{(\ref{eq:1b})--(\ref{eq:1f}),}
\end{align}
\end{subequations}
where, to simplify the notation, we denote the estimated control variable as $\hat{{u}}(t,x)\!=\!\mathcal{U}_\omega(t,x,\hat y)$, while the state variable estimate is denoted as $\hat{{y}}(t, x)\!=\!\mathcal{Y}_\theta(t,x,\hat u)$. 
Given the complexity of solving Problem (\ref{eq:goal}), the goal is to develop a fast and accurate neural PDE optimization surrogate. This approach uses the concept of \emph{proxy optimizers} \citep{kotary2021end}, which is detailed next.

\subsection{Surrogate Controller }
\label{sec:surrogate_controller}
The first objective is to establish a neural network-based mapping that transforms the system state $y(t,x)$ of a PDE-constrained optimization problem \ref{eq:1} to a control action $u(t,x)$.
Practically, this mapping is implemented as a neural network $\mathcal{U}_{\omega}$, which takes the current system state $\hat{y}(t,x)$ and desired terminal state $y(T,x)$ as input, and outputs an array of coefficients $\hat{\bm{c}}(t)$ that are then used to estimate the corresponding optimal decision variables. These coefficients are the weights used to combine a set of basis functions to represent the control action, and are described next.

\textbf{Basis function representation of the control action.}
To approximate the optimal control action using a neural network, we represent the control trajectory ${u}(t, {x})$ as a linear combination of \textit{basis functions}. This representation has also been proposed in \cite{basis_cite_1, basis_cite_2}.
This design choice is motivated by the fact that, although the output of a neural network lies in a finite-dimensional space, combining it with continuous basis functions ${\phi_i(x)}$ allows us to construct a continuous approximation of the control trajectory ${u}(t,{x})$. In this way, the controller model $\mathcal{U}_\omega$ operates in a low-dimensional space, while the basis functions map these finite number of outputs into a continuous-time signal.
As shown in the additional experiments provided in Appendix~\ref{app:exp_direct_u}, direct modeling of continuous-time control actions, either with a neural network or classical method, leads to qualitatively similar performance to our basis expansion approach, but while the computational cost of our method remains extremely low, the running time of the classical methods grows significantly. These results show that our method is suitable for real-time tasks.\\
Consider the control action ${u}(t_k, {x})$, 
where $t_k$ denotes a collocation point; at each time instant $t_k$, the control ${u}(t_k, {x})$ is expressed as a weighted sum of $M$ continuous basis functions $\{\phi_i(x)\}_{i=1}^M$:  
\begin{equation} \label{eq:control_basis}
     {u}(t_k, {x}) = \sum_{i=1}^M c_i(t_k) \, \phi_i(x), \qquad t_k = 0, \ldots, T ,
\end{equation} 
where $\bm c(t_k) = [c_1(t_k) \; \ldots \; c_M(t_k)]^\top$ denotes the coefficients of the basis functions at time instant $t_k$. For example, in one of our experiments, we adopt  sinusoidal basis functions, e.g. $\phi_1(x) = \sin( \pi  x/X)$, $\phi_2(x) = \cos( \pi x/X)$, $\dots$, $\phi_{2M-1}(x) = \sin( M \pi x/X)$, $\phi_{2M}(x) = \cos( M \pi x/X)$, which provide a Fourier-like expansion of the control signal. Here $M$ denotes the number of sinusoidal modes included in the expansion, with higher values of $M$ allowing for better approximations of the control variable.

\textbf{Surrogate controller model training.}
The role of the controller surrogate $\mathcal{U}_\omega$ is to predict the array of coefficients $\bm c(t_k) = [c_1(t_k) \; \dots \; c_M(t_k)]^T$. Formally, the network $\mathcal{U}_\omega$ approximates the function
\begin{equation}
    \label{eq:controller_mapping}
    {U} : \big ({y}(t_k,{x}), y(T,x) \big) \mapsto \bm c(t_k),
\end{equation}
which maps the state variable ${y}(t_k, x)$ and desired terminal state ${y}(T, x)$ to the corresponding controller weights $\bm c (t_k)$, subsequently used to construct the control action $ u(t_k, x)$ according to (\ref{eq:control_basis}). 
The surrogate controller model $\mathcal{U}_\omega$ is trained \textit{on-the-fly}, using a dataset $\mathcal{D}=\{(\hat{{\bm y}}^i(t_k, {x}), \, {y}(T, {x})\}_{i=1}^N$, where $\hat{{\bm y}}(t_k, {x})$ denotes the state variables estimated by the dynamic predictor $\mathcal{Y}_\theta$, ${y}(T, {x})$ is the desired terminal state and $N$ is the number of samples. 
Here, bold symbols $\bm y(t_k,x)$ denote the space-discrete version of the corresponding vector $y(t_k,x)$, with each collocation point corresponding to a specific output of $\mathcal{Y}_\theta$, e.g. $\bm y(t_k,x) = \big[\hat y(t_k,x_1) \; \ldots \; \hat y(t_k,x_n)\big]^T$. 
Note that this setting is \textit{unsupervised}: optimal weights $\bm c(t_k)$ are unknown, and candidate weights $\hat {\bm c}(t_k) = \mathcal{U}_\omega(\hat{\bm y}(t_k,x), y(T,x))$ are evaluated solely by the objective value $\mathcal{J}$ in (\ref{eq:1a}) they induce. 
Figure \ref{fig:DE-OP_scheme} illustrates the interaction between PDE-OP’s proxy optimizer $\mathcal{U}_\omega$ and dynamic predictor $\mathcal{Y}_\theta$ at time instant $t = t_k$.
Given the predicted state $\hat{{y}}(t_{k},{x})$ (for $k=0$ we have $\hat{{y}}(t_{0},{x}) = {y}_0(x)$), the surrogate controller outputs an estimate of $\bm c(t_k)$ which are then used to construct the control action ${u}(t_k, {x})$; this is subsequently used by $\mathcal{Y}_\theta$ to predict the next state $\hat{{y}}(t_{k+1},{x})$. This process is repeated at each collocation point, until the terminal state ${y}(T, {x})$ is estimated. 
\\
To build our surrogate controller, we build on the proxy optimizer method proposed in \citet{park2023self}, which allows self-supervised training without access to precomputed optimal solutions, while directly encouraging feasibility through primal–dual updates.
Once trained, the model ${\cal U}_\omega$ can be used to generate near-optimal solutions at low cost. 
The next section discusses how the state variables are estimated using a discrete-time Neural Operator architecture.

\subsection{Discrete-time Neural Operator} \label{sec: gen-DON}
The second component of PDE-OP is a discrete-time Neural Operator architecture, here denoted as $\mathcal{Y}_\theta$, which models the PDE solution operator. 
The proposed dynamic predictor employs two subnetworks: the first, which we denote \emph{branch network}, encodes input functions (e.g., control signals), and the second, called \emph{trunk network}, encodes the coordinates where the solution is evaluated (e.g., spatial locations). Their outputs are then combined to approximate the target operator. This architecture is inspired by~\citep{Lu_2021}, a neural operator framework designed to approximate mappings between infinite-dimensional function spaces. 

Recall that the PDE-constrained problem couples system dynamics and decision variables. To take account for this coupling, the proposed architecture incorporates both the system state and the controller output. Given the current estimated state $\hat{\bm y}(t_k,x)$ and the control weights $\hat{\bm c}(t_k)$ estimated by $\mathcal{U}_\omega$, the network $\mathcal{Y}_\theta$ outputs an estimate of the state at the subsequent collocation point in time:
\begin{equation} \label{eq:deeponet_update}
    \hat {\bm {y}}(t_{k+1}, x) = \mathcal{Y}_\theta\!\left( t, x, \hat {\bm y}(t_k, x), \hat {\bm{c}}(t_k) \right).
\end{equation}   
Rolling out (\ref{eq:deeponet_update}) across $k=0,\ldots,T-1$ yields the full estimated state trajectory, $\{\hat{\bm y}(t_k, x)\}_{k=0}^T$.


\textbf{Branch network.}  
The branch net encodes the estimated system state sampled at $n$ fixed sensor locations, $\hat {\bm y}(t_k, x)$, 
together with estimated control weights $\hat{\bm c}(t_k)$, and maps this concatenated input to a $d$-dimensional vector of coefficients
$\bm b\!\left(\hat{\bm y}(t_k, x), \hat{\bm c} (t_k)\right) \in \mathbb{R}^d$.\\    
\textbf{Trunk network.}  
The trunk net encodes the spatial query coordinate $x$, producing a latent feature vector $\bm \gamma(x) \in \mathbb{R}^d$. 
\newline The estimated subsequent (in time) state $\hat {\bm  y}(t_{k+1},x)$ at location $x_i$ is obtained by combining the branch coefficients and trunk features through an inner product:  
\begin{equation}
    \hat { y}(t_{k+1},x_i) \;=\; \sum_{j=1}^d b_j\!\left(\hat {{y}}(t_k, x_i), \hat {\bm c}(t_k)\right)\,\gamma_j(x_i), \; \; i=1,\dots,n.
\end{equation} 

The remainder of this section details the method by which the state variables are precisely estimated using the proposed discrete-time Neural Operator.

\textbf{Model initialization and training.} Given the proposed time-discrete neural operator model $\mathcal{Y}_\theta$ takes as input the weights of the basis functions estimated by the surrogate controller $\mathcal{U}_\theta$, it is practical to initialize the dynamic predictor model. 
To achieve this, we initialize $\mathcal{Y}_\theta$ in a supervised fashion; its training data are generated by sampling time-varying control weight sequences $\bm c(t_k)$ from a gaussian distribution, e.g., $c_i(t_k) \sim \mathcal{N}(0,\sigma_i)$, and by specifying the terminal state $y(T,x)$. The corresponding solution $ y(t,x)$ satisfying~(\ref{eq:1b})--(\ref{eq:1f}) is computed using a numerical PDE solver. Each trajectory is discretized into one-step training samples of the form $(y(t_k,x), \bm c(t_k)) \mapsto y(t_{k+1},x)$.  
Formally, the dataset of training pairs is given by  
$\mathcal{D} = \Big\{ \big(y^i(t_k,x), \bm c^i(t_k), y^i(t_{k+1},x)\big) \;\Big|\; 
    t_k \in \{0,\dots,T-1\} \Big\}_{i=1}^N$.
The network $\mathcal{Y}_\theta$ is trained to minimize the prediction error between its outputs and the ground-truth values. Formally, this model is trained by minimizing the loss:  
\begin{equation} \label{eq:deeponet_loss}
    \min_{\theta} \; \mathbb{E}_{(\bm c, y)\sim\mathcal{D}}
    \Big[ \big\| \hat{y}(t, x) - y(t, x) \big\|^2 \Big],
\end{equation}
where $\mathcal{D}$ denotes the dataset of pairs $(\bm c, y)$.
This training strategy ensures that $\mathcal{Y}_\theta$ learns to approximate the PDE solution operator given a distribution of weights $\bm c(t)$ approximating the distribution of estimated weights generated by the surrogate controller $\mathcal{U}_\omega$, so that it can provide accurate PDE-solutions given the predicted controller weights $\hat{\bm c}(t)$. 

\subsection{Handling Static and Dynamics Constraints Jointly}
\label{sec: DE-OP_model}

To integrate the proposed time-discrete neural perator within the decision process, this paper proposes a  Primal Dual (LD) learning approach, which is inspired by the augmented Lagrangian relaxation technique \citep{Hestenes1969MultiplierAG} adopted in classic optimization.
In Lagrangian relaxation, some or all the problem constraints are relaxed into the objective function using Lagrangian multipliers to capture the penalties induced by violating them.
The proposed formulation leverages Lagrangian duality to integrate trainable 
regularization terms that encapsulates violations of the constraint functions. 
When all the constraints are relaxed, the violation-based Lagrangian 
of problem (\ref{eq:1}) is 
\begin{equation}
\label{eq:violation_based_Lagrangian}
\minimize_{{u}} \mathcal{J}( {{u(t,x)}}, {{y}}(t, x)) + \bm {\lambda}^\top_{h^\prime} | \bm h^\prime({{u(t,x)}}, {{y}}(t,x)) | + \bm{\lambda}^\top_g \max(0, \bm g({{u(t,x)}}, {{y}}(t,x))),
\notag
\end{equation}
where $\mathcal{J}$, $\bm g$ are defined in (\ref{eq:1a}), and (\ref{eq:1f}), respectively, and $\bm{h}^\prime$ is defined as follows, 

\begin{equation}
\label{eq:h_prime}
\bm h'(u,y) \;=\;
\begin{bmatrix}
\partial_t y(t,x) - \mathcal{I}_t\!\big(y(t,x),u(t,x),t,x\big) \\[2pt]
\partial_x y(t,x) - \mathcal{I}_x\!\big(y(t,x),u(t,x),t,x\big) \\[2pt]
y(0,x) - y_0(x) \\[2pt]
y(t,x) - y_\partial(t,x) \\[2pt]
\bm h(u,y)
\end{bmatrix}
= \bm 0.
\end{equation}

It denotes the set of \emph{all} equality constraints of problem (\ref{eq:1}), thus extending the constraints $\bm h$ in (\ref{eq:1e}), with the system dynamics (\ref{eq:1b})-(\ref{eq:1c}) and the initial and boundary conditions equations (\ref{eq:1d})-(\ref{eq:1e}) written in an implicit form as above.
Therein, $\bm{\lambda}_{h^\prime}$ and $\bm{\lambda}_g$ are the vectors of Lagrange multipliers associated with functions $\bm{h}^\prime$ and $\bm g$, e.g. ${\lambda}^i_{h^\prime}, {\lambda}^j_g$ are associated with the $i$-th equality $h^\prime_i$ in $\bm{h}^\prime$ and $j$-th inequality $g_j$ in $\bm g$, respectively. The key advantage of expressing the system dynamics (\ref{eq:1b})-(\ref{eq:1c}) and initial and boundary conditions (\ref{eq:1d})-(\ref{eq:1e}) in the same implicit form as the equality constraints $\bm{h}$, (as shown in (\ref{eq:h_prime})), is treating the system dynamics in the same manner as the constraint functions $\bm{h}$. This enables us to satisfy the system dynamics and the static set of constraints ensuring that they are incorporated seamlessly into the optimization process. 
\noindent

The proposed primal-dual learning method uses an iterative approach to find good values of the \emph{primal} $\omega$, $\theta$ and dual $\bm{\lambda}_{h^\prime}, \bm{\lambda}_{g}$ variables; it uses an augmented modified Lagrangian as a loss function to train the prediction $\hat{{u}}$, $\hat{{y}}(t)$ as employed 
\begin{equation}
\label{eq:LD_loss}
\begin{split}
\mathcal{L}^{\text{PDE-OP}}(\hat{{u}},\hat{{y}}) &\!\!=\!\! \mathcal{J}( {{\hat u(t,x)}}, {{\hat y}}(t, x)) \!+\! \bm {\lambda}^\top_{h^\prime} | \bm h^\prime(\hat{{u}}(t,x), \hat{{y}}(t,x)) | \! + \! \bm{\lambda}^\top_g \max(0, \bm g(\hat{{u}}(t,x), \hat{{y}}(t,x) )),\!\!\!\!
\end{split}
\end{equation}
where 
$\mathcal{J}( {{\hat u(t,x)}}, {{\hat y}}(t, x))$ represents the total objective cost, while $\bm {\lambda}^\top_{h^\prime} | \bm h^\prime(\hat{{u}}(t,x), \hat{{y}}(t,x) ) |$ and $\bm{\lambda}^\top_g \max(0, \bm g(\hat{{u}}(t,x), \hat{{y}}(t, x)))$ measures the constraint violations incurred by prediction $\hat{{u}}(t,x)$ and $\hat{{y}}(t,x)$. 
The loss function in (\ref{eq:LD_loss}) combines an objective term to ensure minimization of the objective function (\ref{eq:1a}) with a weighted penalty on violations of the constraint functions  $\bm{h}^\prime, \bm{g}$. This accounts for the contribution of both networks $\mathcal{U}, \mathcal{Y}$ during training, which is balanced via iterative updates of the  multipliers in (\ref{eq:LD_loss}) based on the amount of violation of the associated constraint function.

At iteration $j+1$, finding the optimal parameters $\omega, \theta$ requires solving 
\begin{equation}
\omega^{j+1}, \theta^{j+1} = \arg\min_{\omega, \theta} \mathbb{E}_{y(t,x) \sim \mathcal{D}} \left[ \mathcal{L}^{\text{PDE-OP}} \left( \mathcal{U}_{\omega^j}^{\bm \lambda^j}(t, x, \hat y),  \mathcal{Y}_{\theta^j}^{\bm \lambda^j}\left( t,x,\hat u \right) \right) \right],
\end{equation}
where $\mathcal{U}_{\omega}^{\lambda^j}$ and $ \mathcal{Y}_{\theta}^{\lambda^j}$ denote the DE-OP's optimization and 
predictor models $\mathcal{U}_{\omega}$ and $\mathcal{Y}_{\theta}$, at iteration $j$, with $\bm \lambda^j = [\bm {\lambda}^j_{h^\prime} \ \bm {\lambda}^j_{g}]^\top$. This step is approximated using a stochastic gradient descent method
\begin{equation}
\begin{split}
\omega^{j+1} &= \omega^j - \eta \nabla_{\omega} \mathcal{L}^{\text{PDE-OP}}\left( \mathcal{U}_{\omega^j}^{\bm \lambda^j}(t, x, \hat y),  \mathcal{Y}_{\theta^j}^{\bm \lambda^j}\left( t,x,\hat u \right) \right) \\
\theta^{j+1} &= \theta^j - \eta \nabla_{\theta} \mathcal{L}^{\text{PDE-OP}}\left( \mathcal{U}_{\omega^j}^{\bm \lambda^j}(t, x, \hat y),  \mathcal{Y}_{\theta^j}^{\bm \lambda^j}\left( t,x,\hat u \right) \right)
\end{split}
\end{equation}
where $\eta$ denotes the learning rate and $\nabla_{\omega} \mathcal{L}$ and  $\nabla_{\theta} \mathcal{L}$ represent the gradients of the loss function $\mathcal{L}$ with respect to the parameters $\omega$ and $\theta$, respectively, at the current iteration $k$. Importantly, this step does not recompute the training parameters from scratch, but updates the weights $\omega$, $\theta$ based on their value at the previous iteration.
Finally, the Lagrange multipliers are updated as
\begin{equation}
\begin{split}
\bm {\lambda}^{j+1}_{h^\prime} &=  \bm {\lambda}^{j}_{h^\prime} + \rho | \bm h^\prime(\hat{{u}}(t,x), \hat{{y}}(t, x)) | \\
\bm {\lambda}^{j+1}_{g} &=  \bm {\lambda}^{j}_{g} + \rho \max \left(\bm 0, \bm g(\hat{{u}}(t,x), \hat{{y}}(t, x)) \right )
\end{split}
\end{equation}
where $\rho$ denotes the Lagrange step size and the $\text{max}$ operation is performed element-wise.

Next, we evaluate our method on a range of PDE-constrained optimization tasks with varying levels of complexity, including cases space-time varying control actions and nonlinear PDE dynamics.

\section{Experimental Setting}
\label{sec:experiments}

In this section, we evaluate the proposed method on three optimal control tasks of increasing complexity. In these tasks the dynamics are described by the following PDE equations: Voltage Dynamics (linear PDE with time-invariant control input $u(x)$), 1D Heat equation (linear PDE with time-variant control input $u(t, x)$), and 1D viscous Burgers’ equation (nonlinear PDE with time-variant control input $u(t, x)$). PDE-OP is compared against state-of-the-art approaches for optimal control problems: the Direct method~\citep{Direct-Method}, the Adjoint-Sensitivity method~\citep{adjoint-method}, and (non-linear) Model Predictive Control (NMPC)~\citep{MPC}. These methods are widely adopted in the literature, all of which rely on numerical PDE solvers whenever dynamics or gradients are evaluated. The Open-loop (direct/adjoint) Control method requires a full forward rollout per gradient step and a backward adjoint evalution, while MPC solves a finite-horizon problem at every control step (i.e., a receding horizon). The runtime of these methods is determined by repeated calls to the underlying numerical PDE solver and grows with the spatial grid size ($n$), horizon length ($T$), and complexity (e.g., non-linearity) of the PDE dynamics. For each experiment and method, we report the accuracy, measured as the $L_2$ error between the terminal state and the predicted terminal state, $\| y(T,x) - \hat{y}(T,x) \|^2_2$, the inference time of PDE-OP and runtime of the classical methods, both expressed in seconds. For each task, we assess the capability of the proposed time-discrete neural operator to capture system dynamics by comparing its outputs against the solutions obtained from a numerical PDE solver.
\\
Please refer to Appendices~\ref{app:classical_methods}, and \ref{app:hyperparam_settings} for details on the baseline methods and hyperparameter settings, respectively.
In Appendix~\ref{app:dtno_direct_u} we report an ablation to evaluate the impact of our basis functions, introduced in Section \ref{sec:surrogate_controller}). In summary, this experiment shows that 
using a full formulation is impractical for NMPC, pushing solve times well beyond real-time limits. Finally, we show that Adjoint methods suffer from inaccurate gradients, which leads to a marked decrease in solution quality.

\subsection{Voltage Control Optimization} \label{exp:voltage}

The problem of regulating the voltage magnitude in a leaky electrical transmission line, subject to a reaction-diffusion PDE with Neumann conditions is described as follows:
\begin{subequations}
\label{eq:rd_problem}
\begin{alignat}{2}
    \minimize_{u(x)} \quad & \int_{\Omega} (y(T, x) - y_{\text{target}}(x))^2 dx + \gamma \int_{\Omega} u(x)^2 dx \label{rd:obj} \\
    \text{s.t.} \quad & \frac{\partial y}{\partial t} = D \frac{\partial^2 y}{\partial x^2} - \beta(y - y_{\text{ref}}(x)) + \alpha u(x), \quad && \forall (x, t) \in \Omega \times [0, T] \label{rd:pde} \\
    & y(0, x) = y_0(x), \quad && \forall x \in \Omega,  \{\ t = 0\} \label{rd:pde_init} \\
    & \frac{\partial y}{\partial x}(t, x) = 0, \quad && \forall x \in \partial\Omega, \ t \in [0, T] \label{rd:pde_bc} \\
    & u_{\text{min}} \leq u(x) \leq u_{\text{max}}. \quad && \forall x \in \Omega. \label{rd:ineq}
\end{alignat}
\end{subequations}
Here, $y(t, x) \in \mathbb{R}$ represents the voltage profile across the space-time domain defined by the cartesian product between $\Omega$ and time horizon $[0, T]$. The system dynamics, expressed by~(\ref{rd:pde}), model the physical phenomena of diffusion ($D \frac{\partial^2 y}{\partial x^2}$), voltage leakage towards a reference $y_{\text{ref}}(x)$ ($-\beta(y - y_{\text{ref}}(x))$) 
and the applied control action ($\alpha u(x)$). $D$, $\beta$, and $\alpha$ are all constants representing the diffusion rate, leakage rate, and control gain, respectively.
The system is initialized from a state $y_0(x)$ according to~(\ref{rd:pde_init}) and is constrained by zero-flux Neumann boundary conditions~(\ref{rd:pde_bc}), implying no voltage diffusion across the boundaries of the domain. The control decision $u(x)$ is a space-dependent function representing the externally applied voltage, constrained by actuator limits in~(\ref{rd:ineq}). The objective~(\ref{rd:obj}) minimizes a combination of terminal cost and control energy.

\textbf{Datasets and methods.} In the system dynamics~(\ref{rd:pde}), we set the diffusion rate $D=0.1$, leakage rate $\beta=1.0$, and control gain $\alpha=2.0$, with a uniform reference voltage $y_{\text{ref}}(x)=1.0$. The ground-truth dynamics are simulated using a second-order implicit Crank-Nicolson finite-difference method\citep{Crank_Nicolson_1947}. 
To initialize the proposed time-discrete Neural Operator $\cal Y_\theta$, we generate a dataset by sampling a diverse set of control functions $u(x)$ from a zero-mean Gaussian Random Field (GRF) with a squared-exponential kernel, with mean $m(x)$ and covariance function $k_{\ell}(x_1,x_2)$ as:
\begin{equation}
    m(x)= 0,\qquad
    k_{\ell}(x_1,x_2) \;=\; \sigma^2 \exp\!\left(-\frac{\|x_1-x_2\|_2^2}{2\ell^2}\right),
\end{equation}
where $\ell>0$ is the length–scale and $\sigma^2>0$ the marginal variance. The corresponding solution $y(t, x)$ is computed using a numerical solver,
and then split into $80 \%$ training, $10\%$ validation, and $10\%$ test set.  

PDE-OP's model $\mathcal{Y}_\theta$ 
estimates the voltage profile $y(t, x)$; for $t=T$ this prediction yield $y(T, x)$, which is compared against the target profile $y_{\text{target}}(x)$. PDE-OP's surrogate controller $\mathcal{U}_\omega$ is trained to produce optimal control action $\hat{u}(x)$ 
while minimizing objective~(\ref{rd:obj}). Our approach is benchmarked against Direct and Adjoint-Method, which are described in Appendix \ref{app:dm_baseline}- \ref{app:adj_baseline}. 
At inference time, PDE-OP's surrogate controller $\mathcal{U}_\omega$ is tested on three unseen target profiles of different level of complexities: constant ($y_{\text{target}}(x) = p_1$), linear ramp ($y_{\text{target}}(x)= p_1x+p_2$), and sine wave ($y_{\text{target}}(x)= p_1 + p_2(\sin f_0x+p_3)$). For each target, the control $\hat{u}(x)$ is generated in a single forward pass.



\begin{figure}[t!]
\centering
\includegraphics[width=.92\linewidth]{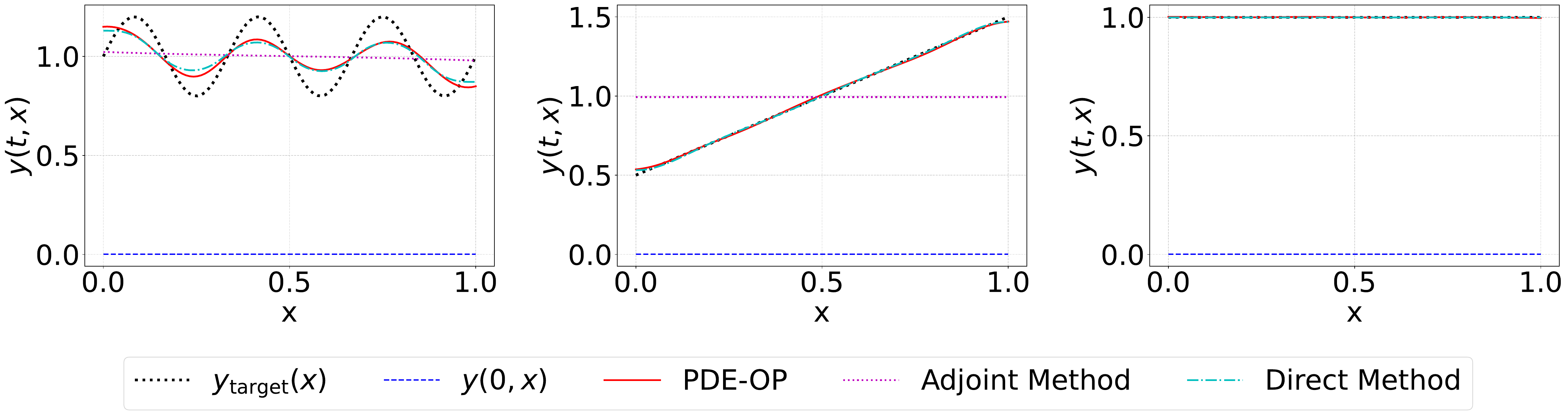}
\caption{Voltage Control Optimization; comparison of PDE-OP and classical methods for three target profiles: $y_{\text{target}}(x)=1+0.2\sin(6x)$ (left), $y_{\text{target}}(x)=x+0.5$ (center), and $y_{\text{target}}(x)=1$ (right)}
\label{fig:Comparison for Voltage}
\end{figure}
\begin{figure}[t!]
\centering
\includegraphics[width=0.95\linewidth]{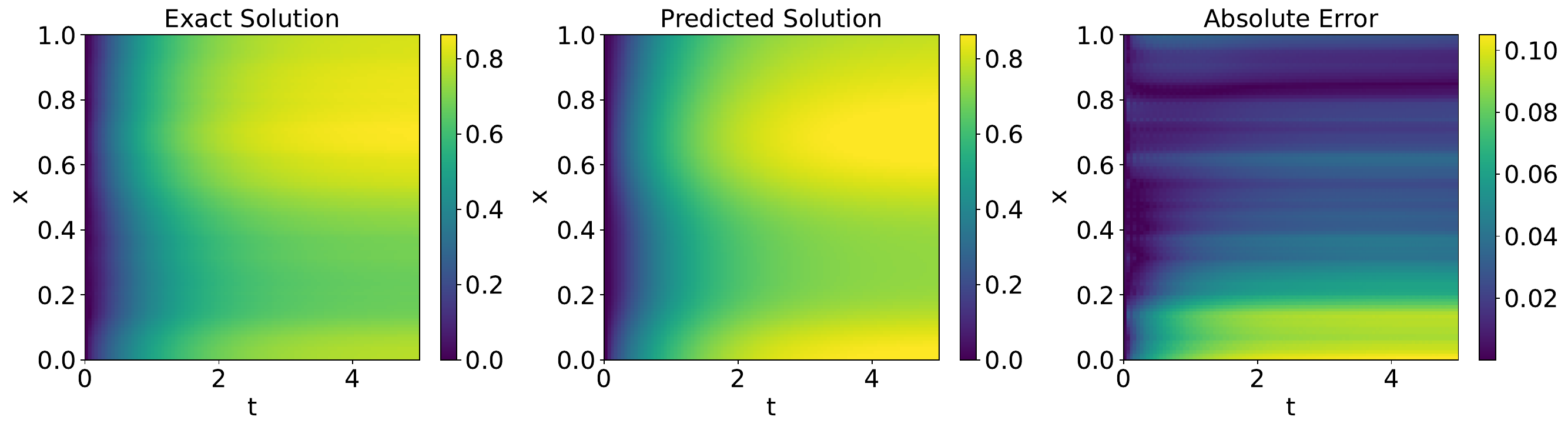}
\caption{Voltage control optimization; comparison between the solution estimate (center) of PDE-OP's dynamic predictor $\cal Y_\theta$ at test time with the solution obtained with a numerical algorithm (left) and related absolute error (right).}
\label{fig:Exact vs predicted on voltage}
\end{figure}

\noindent\textbf{Results.}  
Figure \ref{fig:Exact vs predicted on voltage} shows the predicted voltage (central sub-figure of Fig. \ref{fig:Exact vs predicted on voltage}) of PDE-OP's dynamic component $\mathcal{Y}_\theta$ at test time, which is compared to the solution (left sub-figure) obtained with a numerical PDE-solver. 
The figure shows that the predicted voltage closely matches the numerical PDE-solver solution, with an absolute error (right sub-figure) on the order of $10^{-2}$ across the space–time domain.


Figure \ref{fig:Comparison for Voltage}  presents a comparison between PDE-OP's predicted terminal state $\hat{y}(T,x)$ and that obtained using the classical methods, over $3$ target profiles $y_{\text{target}}(x)$: sinusoidal, linear and constant.
This figure shows that the voltage distribution produced by PDE-OP is well aligned with those produced by the Direct and the Adjoint-sensitivity method. 
As shown in the first row of Table \ref{tab:exp_summary}, the direct method is the slowest, requiring $26.097 s$ and yielding an MSE of $7.46 \times 10^{-3}$. 
The Adjoint method runs in $0.755 s$ 
with a similar accuracy, $7.13\times 10^{-3}$. 
Notably, \textsc{PDE-OP} is the fastest, by a significant margin: it runs in only $0.035$ s ($\sim\!21.6\times$ faster than Adjoint; $\sim\!746\times$ faster than Direct) and yield the lowest MSE, with $6.95\times 10^{-3}$.
These results show the enormous potential of the proposed approach, which produces solutions that are qualitatively similar (or even slightly better) 
than the classical methods, while being between $20$ and $750$ times faster than them.

\begin{table}[h!]
\centering
\setlength{\tabcolsep}{4.0pt} 
\renewcommand{\arraystretch}{1.0} 
\caption{Comparison across PDE tasks and classical  methods. 
All results use sinusoidal target profiles. Further details and results for linear and constant profiles can be found in Appendix~\ref{app:all_experiments}.} 
\label{tab:exp_summary}
\begin{tabular}{@{}lllll@{}}
\toprule
\textbf{PDE System} & \textbf{Control Type} & \textbf{Methods} & \textbf{Runtime} & \textbf{Accuracy} \\ \midrule
\begin{tabular}[c]{@{}l@{}}\textbf{Voltage Control}\\(Reaction-Diffusion)\end{tabular} &
Open-Loop &
\begin{tabular}[c]{@{}l@{}}Direct (Finite Diff.)\\Adjoint-Method\\\textbf{PDE-OP (ours)}\end{tabular} &
\begin{tabular}[c]{@{}l@{}}26.097\\0.755\\\textbf{0.035}\end{tabular} &
\begin{tabular}[c]{@{}l@{}}0.00746\\0.00713\\\textbf{0.00695}\end{tabular} \\ \midrule
\begin{tabular}[c]{@{}l@{}}\textbf{1D Heat Equation}\\(Linear PDE)\end{tabular} &
Closed-Loop &
\begin{tabular}[c]{@{}l@{}}LMPC\\Adjoint-Method\\\textbf{PDE-OP (ours)}\end{tabular} &
\begin{tabular}[c]{@{}l@{}}0.384\\0.662\\\textbf{0.124}\end{tabular} &
\begin{tabular}[c]{@{}l@{}}\textbf{0.00035}\\\textbf{0.00035}\\\textbf{0.00035}\end{tabular} \\ \midrule
\begin{tabular}[c]{@{}l@{}}\textbf{1D Burgers' Equation}\\(Nonlinear PDE)\end{tabular} &
Closed-Loop &
\begin{tabular}[c]{@{}l@{}}NMPC\\Adjoint-Method\\\textbf{PDE-OP (ours)}\end{tabular} &
\begin{tabular}[c]{@{}l@{}}878.667\\120.696\\\textbf{0.062}\end{tabular} &
\begin{tabular}[c]{@{}l@{}}\textbf{0.00007}\\0.00023\\0.00030\end{tabular} \\
\bottomrule
\end{tabular}
\end{table}

\begin{figure}[t!]
\centering
\includegraphics[width=.92\linewidth]{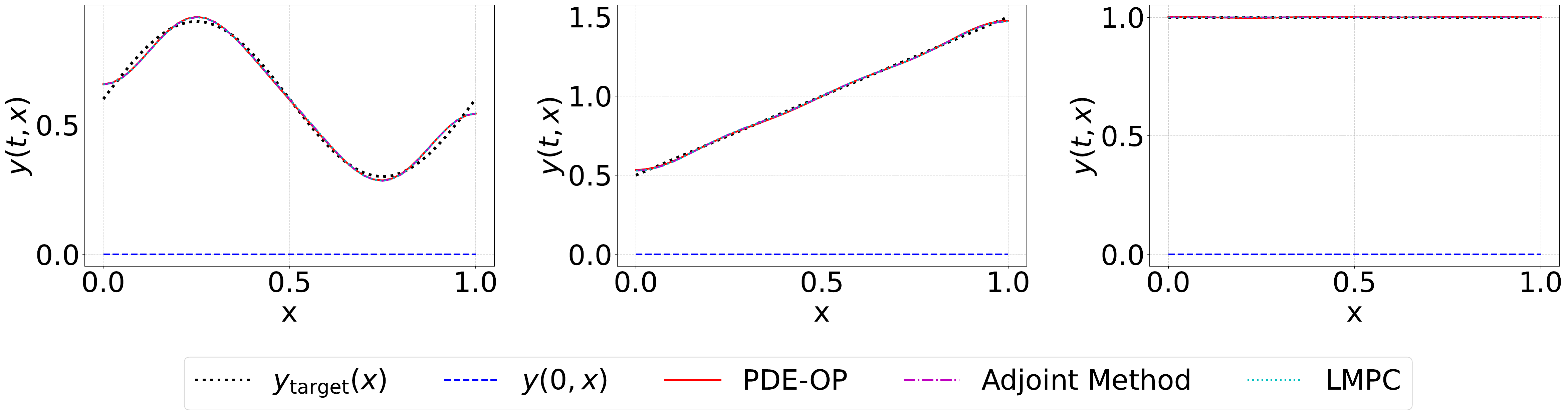}
\vspace{-6pt}
\caption{Optimal Control of the Heat Equation; comparison of PDE-OP and classical methods for three target profiles: $y_{\text{target}}(x) = 0.6 +0.3(\sin(2x))$,  $y_{\text{target}}(x) = x+0.5$, and $y_{\text{target}}(x) = 1$.}
\label{fig:Comparison_heat_weights}
\end{figure}

\begin{figure}[t!]
\centering
\includegraphics[width=.95\linewidth]{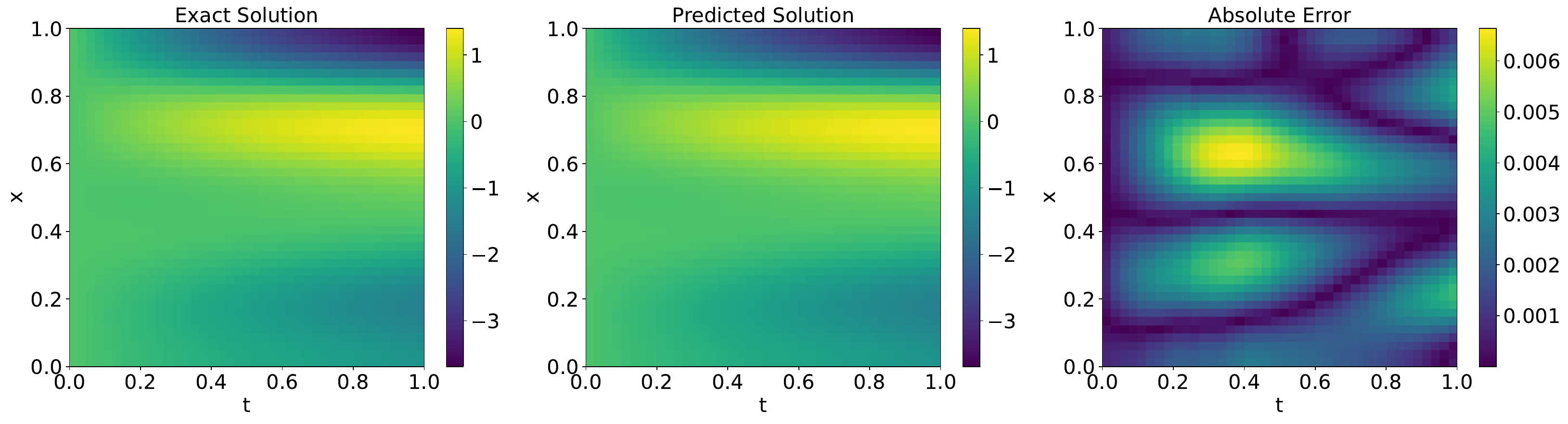}
\caption{Optimal Control of the 1D Heat Equation: comparison between the solution estimate (center) of PDE-OP's dynamic predictor $\cal Y_\theta$ at test time with the solution obtained with a numerical algorithm (left) and related absolute error (right).}
\label{fig:Exact vs predicted on Heat}
\end{figure}

\subsection{Optimal Control of the Heat Equation} \label{sec:Heat}

This task involves controlling the evolution of temperature in a one-dimensional domain described by a reaction–diffusion PDE. Given an initial profile, the controller aims to reach a prescribed target state with minimal control effort. The problem is formulated as:

\begin{subequations}
\label{eq:heat_problem}
\begin{alignat}{2}
    \minimize_{\bm c(t)_{t \in [0,T]}} \quad & \mathcal{L} = \mathcal{L}_{\text{terminal}} + \lambda \mathcal{L}_{\text{running}} + \gamma \mathcal{L}_{\text{effort}} \label{heat:obj} \\
    \text{s.t.} \quad & \frac{\partial y}{\partial t} = D \frac{\partial^2 y}{\partial x^2} - \beta(y - y_{\text{ref}}(x)) + \alpha u(t,x), \quad && \forall (x, t) \in \Omega \times [0, T] \label{heat:pde} \\
    & u(t, x) = \bm c(t)^\top \bm{\phi}(x), \quad && \forall (x, t) \in \Omega \times [0, T] \label{heat:control_rep} \\
    & y(0, x) = y_0(x), \quad && \forall x \in \Omega \label{heat:pde_init} \\
    & \frac{\partial y}{\partial x}(t, x) = 0, \quad && \forall x \in \partial\Omega, \ t \in [0, T] \label{heat:pde_bc} \\
    & c_{\text{min}} \leq \bm c(t) \leq c_{\text{max}}, \quad && \forall \ t \in [0, T]. \label{heat:ineq}
\end{alignat}
\end{subequations}
The loss components in (\ref{heat:obj}) defines as follows.
\begin{subequations}
\label{eq:heat_losses_integral}
\begin{align}
    \mathcal{L}_{\text{terminal}} &:= \int_{\Omega} \left| y(T, x) - y_{\text{target}}(x) \right|^2 \,dx, \label{heat:loss_term_int} \\
    \mathcal{L}_{\text{running}} &:= \int_{0}^{T} \int_{\Omega} \left| y(t, x) - y_{\text{target}}(x) \right|^2 \,dxdt, \label{heat:loss_run_int} \\
    \mathcal{L}_{\text{effort}} &:= \int_{0}^{T} ||\bm c(t)||_2^2 \,dt. \label{heat:loss_eff_int}
\end{align}
\end{subequations}
Here $y(t,x)$ represents the temperature profile. 
The objective (\ref{heat:obj}) balances three terms: a terminal loss for final state accuracy, a running loss for temperature tracking, and an effort loss for control efficiency, described by (\ref{heat:loss_term_int})--(\ref{heat:loss_eff_int}). 

\textbf{Datasets and methods.} The solutions of (\ref{heat:pde}) are generated using a Crank-Nicolson method, which deal with time-varying source term by averaging the control input over each time step. To train our models, we generate a dataset of  diverse trajectories. First, smooth, time-varying control weights ($\bm c(t)$) are generated by sampling each of the $M$ weight trajectories from a GRF over the time domain. The full spatial-temporal control field $u(t, x)$ is then reconstructed from these weights constrained by (\ref{heat:ineq}) using a set of fixed basis functions, as defined in (\ref{heat:control_rep}). For each generated control field, the full PDE is solved to generate a ground-truth temperature evolution $y(t, x)$.
Each complete simulation yields a training sample, consisting of the control input weight trajectory $\{ \bm c(t_k)\}$, and the corresponding state trajectory $\{  y(t_k, x)\}$ evaluated at a set of fixed sensor locations. The dataset is then split into 80\% training, 10\% validation and 10\% test.

Our framework uses two neural networks. First discrete-time neural operator, $\mathcal{Y}_\theta$, is trained as a surrogate model described in Section (\ref{sec: gen-DON}). It act as a one-step time-advancement operator, learning the mapping $(y(t_k, x), \bm c(t_k)) \rightarrow y(t_{k+1}, x)$. Second, PDE-OP's surrogate controller $\mathcal{U}_\omega$ is a recurrent neural network that acts as sequential decision-maker. At each time step $t_k$, it takes the current state $y(t_k, x)$ and the final target $y_{\text{target}}(x)$ to produce the control weights $(\bm c(t_k))$. The controller is trained end-to-end by unrolling the trajectory using the surrogate $\mathcal{Y}_\theta$. We benchmark our approach against a highly-tuned Adjoint-Method and Linear MPC(LMPC). Further implementation details are in Appendix \ref{app:tv_adj} - \ref{app:mpc}.
 
At inference, we evaluate PDE-OP model on three unseen target profiles: constant($y_{\text{target}}(x) = p_1$), linear ramp($y_{\text{target}}(x)= p_1x+p_2$), and sine wave($y_{\text{target}}(x)= p_1 + p_2(\sin f_0x+p_3)$).

\noindent\textbf{Results.}  
Figure \ref{fig:Exact vs predicted on Heat} shows the temperature (central sub-figure of Fig. \ref{fig:Exact vs predicted on Heat}) estimated by PDE-OP's dynamic component $\mathcal{Y}_\theta$ at test time, which is compared to the solution (left sub-figure) obtained with a numerical PDE-solver. The figure shows that also in this setting the predicted temperature aligns closely with the numerical PDE-solver solution, with an absolute error on the order of $10^{-3}$ across the space–time domain.

Figure \ref{fig:Comparison_heat_weights} illustrates the comparison of PDE-OP 's predicted terminal state $\hat{y}(T,x)$ and those obtained using the classical methods, over the same target profiles introduced in the previous experiment.
As also shown in Table \ref{tab:exp_summary}, all three methods achieve the same accuracy, reporting a MSE of $3.5\times 10^{-4}$. However, \textsc{PDE-OP} is by far the fastest method, running in $0.124\,\text{s}$. This corresponds to a $3.1\times$ speedup over LMPC (running in $0.384\,\text{s}$) and a $5.3\times$ speedup over the Adjoint method (which runs in $0.662\,\text{s}$).
These results demonstrate that even in a more complex setting, with time-variant control actions and linear PDE-dynamics, our method produces solutions that are qualitatively comparable to the classical methods, but at a significantly reduced computation cost.

    
    



\subsection{Optimal Control of the 1D Burgers' Equation} \label{sec:burger}

To test the ability of PDE-OP to handle nonlinear dynamics, we address the control of the 1D viscous Burgers' equation with Dirichlet boundary condition. The control input and external force $u(t, x)$, is parameterized identically to the heat equation task, using time-varying weights $(\bm c(t_k))$ for a set of basis functions. The optimization goal is to find the optimal trajectory of these weights subject to Burger's equation (\ref{burgers:pde}), initial (\ref{burgers:pde_init}) and Dirchlet's boundary condition (\ref{burgers:pde_bc}) and box constraints (\ref{burger:ineq}).

\begin{subequations}
\label{eq:burgers_problem}
\begin{align}
    \minimize_{\bm c(t)_{t \in [0,T]}} \quad & \mathcal{L} = \mathcal{L}_{\text{terminal}} + \lambda \mathcal{L}_{\text{running}} + \gamma \mathcal{L}_{\text{effort}} \label{burgers:obj} \\
    \text{s.t.} \quad & \frac{\partial y}{\partial t} + y \frac{\partial y}{\partial x} = \nu \frac{\partial^2 y}{\partial x^2} + u(t, x), && \forall (x, t) \in \Omega \times [0, T] \label{burgers:pde} \\
    & u(t, x) = \bm c(t)^\top \bm{\phi}(x), && \forall (x, t) \in \Omega \times [0, T] \label{burgers:control_rep} \\
    & y(0, x) = y_0(x), && \forall x \in \Omega \label{burgers:pde_init} \\
    & y(t, x) = 0, && \forall x \in \partial\Omega, \ t \in [0, T] \label{burgers:pde_bc} \\
    & c_{\text{min}} \leq \bm c(t) \leq c_{\text{max}}, \quad && \forall \ t \in [0, T]. \label{burger:ineq}
\end{align}
\end{subequations}
$y(t, x)$ is velocity field, and $\nu$ represents the viscosity parameter. The PDE in (\ref{burgers:pde}) contains convective nonlinearity $(y \frac{\partial y}{\partial x})$ which has a significant control challenge not present in the linear heat equation. The objective retrains the same structure, balancing final state accuracy, path tracking, and control efficiency.

\textbf{Datasets and methods.} The ground-truth dynamics for the nonlinear Burgers' equation are simulated using a high-fidelity numerical solver. The dataset generation process mirrors that of the heat equation: time-varying weights $\bm c(t)$ are sampled from a GRF, and the force field $u(t, x)$ is reconstructed using spatial basis functions, and the PDE is solved to generate trajectories.
Due to the system's high nonlinearity, we benchmark our method against a much more general and computationally expensive Nonlinear MPC (NMPC) and Adjoint-Sensitivity Method. Further implementation details are in Appendix \ref{app:tv_adj} - \ref{app:mpc}.

At inference, we test our PDE-OP model on three unseen target profiles: constant ($y_{\text{target}}(x) = p_1$), parabolic ($y_{\text{target}}(x) = p_1x(1-x)$) and sine wave ($y_{\text{target}}(x)= p_1(\sin f_0x+p_2)$). We don't use a a linear target profile, because of the Dirichlet BCs.


\begin{figure}[t]
\centering
\includegraphics[width=.92\linewidth]{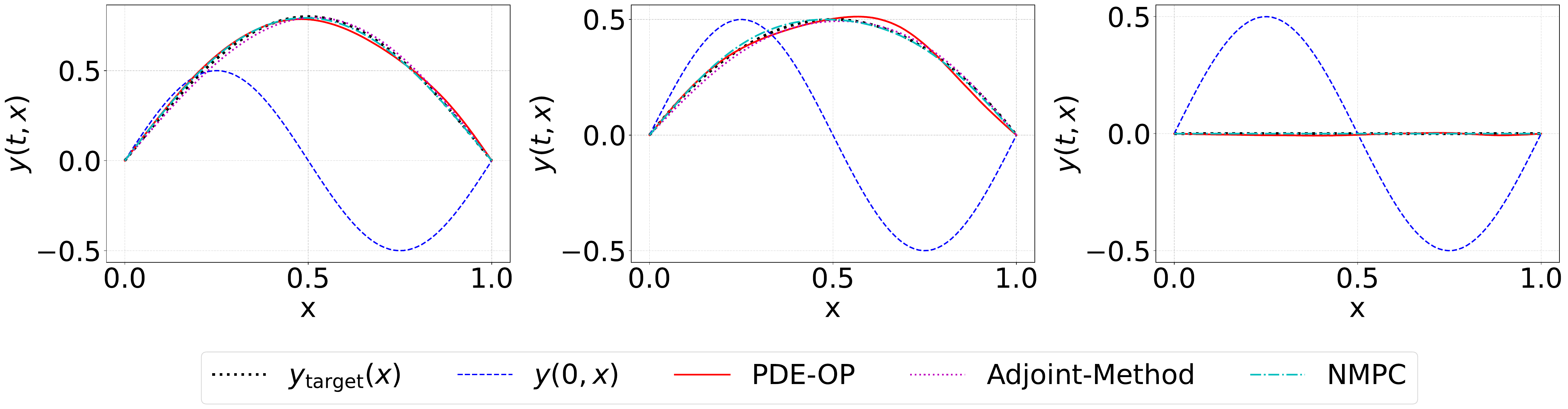}
\vspace{-6pt}
\caption{Optimal control of the 1D Burgers' equation; comparison of PDE-OP and classical methods for three target profiles: $y_{\text{target}}(x) = 0.8(\sin(x))$ (left),  $y_{\text{target}}(x) = 2x(1-x)$ (center), and $y_{\text{target}}(x) = 0$ (right).}
\label{fig:Comparison_burgers_weights}
\end{figure}

\begin{figure}[t!]
\centering
\includegraphics[width=.95\linewidth]{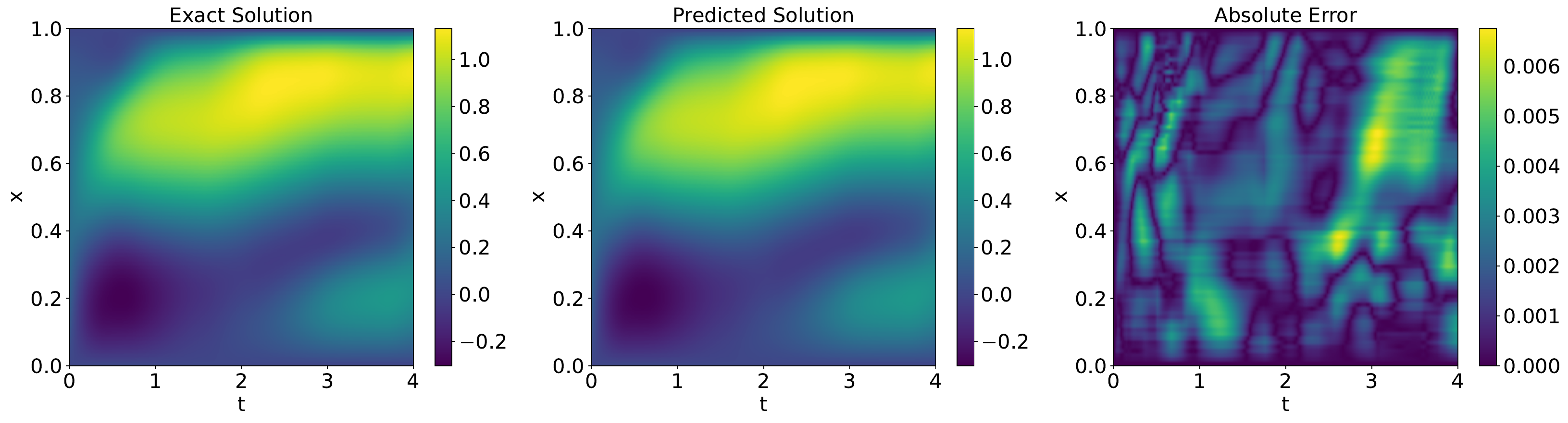}
\caption{Optimal control of the 1D Burgers' equation: comparison between the solution estimate (center) of PDE-OP's dynamic predictor $\cal Y_\theta$ at test time with the solution obtained with a numerical algorithm (left) and related absolute error (right).}
\label{fig:Exact vs predicted Burger}
\end{figure}

\noindent\textbf{Results.}
Figure \ref{fig:Exact vs predicted Burger} shows the system dynamics (central sub-figure of Fig. \ref{fig:Exact vs predicted Burger}) estimated by PDE-OP's dynamic component $\mathcal{Y}_\theta$ at test time, which is compared to the solution (left sub-figure) obtained with a numerical PDE-solver. The figure shows that the predicted temperature aligns closely with the numerical PDE-solver solution; with absolute error on the order of $10^{-3}$ across the space–time domain, demonstrating the ability of PDE-OP's dynamic predictor to accurately capture system dynamics even under nonlinearities.

Figure \ref{fig:Comparison_burgers_weights} illustrates the comparison of PDE-OP 's predicted terminal state $\hat{y}(T,x)$ and those obtained using the classical methods, over the same target profiles introduced in the previous experiments.
As shown in Figure~\ref{fig:Comparison_burgers_weights} and Table~\ref{tab:exp_summary}, each method produces highly accurate solutions, with NMPC achieving the lowest error ($\text{MSE}=7.0\times10^{-5}$) but requiring $878.667 s$. The Adjoint method is slightly less accurate, yielding a 
$\text{MSE}=2.3\times10^{-4}$ in $120.696 s$. In contrast, \textsc{PDE-OP} produces qualitatively comparable predictions while being the fastest $0.062 s$ by a big margin - \textit{approximately
$1.42\times10^{4}$ faster than NMPC and $1.95\times10^{3}$ faster than the Adjoint method!}


\section{Conclusion}\label{sec: conclude}

This work was motivated by the efficiency requirements associated with solving partial differential equations-constrained optimization problems. 
It introduced a novel learning-based framework, PDE-OP, which incorporates differential equation constraints into optimization tasks for real-time applications. The approach uses a dual-network architecture, with one approximating the control strategies and another capturing the associated PDEs. This architecture exploits a primal-dual method to ensure that both the dynamics dictated by the PDEs and the optimization objectives are concurrently learned and respected. This integration enables end-to-end differentiation, allowing for efficient gradient-based optimization, and, to our knowledge, solves PDE-constrained optimization problems in real-time for the first time.
Empirical evaluations across a set of PDE-constrained optimization tasks illustrated PDE-OP’s capability to address these complex challenges adeptly. Our comprehensive results demonstrate the effectiveness of our approach and its broad potential applicability across various scientific and engineering domains where system dynamics arise in optimization or control processes.

\section{Acknowledgments}

The material is based upon work supported by National Science Foundations (NSF) awards 2533631, 2401285, 2334448, and 2334936, and Defense Advanced Research Projects Agency (DARPA) under Contract No.~\#HR0011252E005. 
Any opinions, findings, conclusions, or recommendations expressed in this material are those of the authors and do not necessarily reflect the views of NSF or DARPA.

\bibliography{iclr2026_conference}

@article{willcox2021physics,
  title={The imperative of physics-based machine learning and the role of reduced-order models},
  author={Willcox, Karen and Ghattas, Omar},
  journal={Nature Computational Science},
  volume={1},
  number={3},
  pages={166--168},
  year={2021},
  publisher={Nature Publishing Group}
}

@misc{raissi2017physics,
      title={Physics Informed Deep Learning (Part I): Data-driven Solutions of Nonlinear Partial Differential Equations}, 
      author={Maziar Raissi and Paris Perdikaris and George Em Karniadakis},
      year={2017},
      eprint={1711.10561},
      archivePrefix={arXiv},
      primaryClass={cs.AI}
}

@misc{li2021fourier,
      title={Fourier Neural Operator for Parametric Partial Differential Equations}, 
      author={Zongyi Li and Nikola Kovachki and Kamyar Azizzadenesheli and Burigede Liu and Kaushik Bhattacharya and Andrew Stuart and Anima Anandkumar},
      year={2021},
      eprint={2010.08895},
      archivePrefix={arXiv},
      primaryClass={cs.LG}
}

@article{Hestenes1969MultiplierAG,
  title={Multiplier and gradient methods},
  author={Magnus R. Hestenes},
  journal={Journal of Optimization Theory and Applications},
  year={1969},
  volume={4},
  pages={303-320},
}

@misc{kotary2024learningjointmodelsprediction,
      title={Learning Joint Models of Prediction and Optimization}, 
      author={James Kotary and Vincenzo Di Vito and Jacob Cristopher and Pascal Van Hentenryck and Ferdinando Fioretto},
      year={2024},
      eprint={2409.04898},
      archivePrefix={arXiv},
      primaryClass={cs.LG},
      url={https://arxiv.org/abs/2409.04898}, 
}

@article{mayne2000constrained,
  title={Constrained model predictive control: Stability and optimality},
  author={Mayne, David Q and Rawlings, James B and Rao, CV and Scokaert, PO},
  journal={Automatica},
  volume={36},
  number={6},
  pages={789--814},
  year={2000},
  publisher={Elsevier}
}

@inproceedings{donti2020dc3,
title={DC3: A learning method for optimization with hard constraints},
author={Donti, Priya L and Rolnick, David and Kolter, J Zico},
booktitle={ICLR},
year={2020}
}

@inproceedings{fioretto2020lagrangian,
  title={Lagrangian duality for constrained deep learning},
  author={Fioretto, Ferdinando and Hentenryck, Pascal Van and Mak, Terrence WK and Tran, Cuong and Baldo, Federico and Lombardi, Michele},
  booktitle={Joint European Conference on Machine Learning and Knowledge Discovery in Databases},
  pages={118--135},
  year={2020},
  organization={Springer}
}

@misc{kovachki2024neuraloperatorlearningmaps,
      title={Neural Operator: Learning Maps Between Function Spaces}, 
      author={Nikola Kovachki and Zongyi Li and Burigede Liu and Kamyar Azizzadenesheli and Kaushik Bhattacharya and Andrew Stuart and Anima Anandkumar},
      year={2024},
      eprint={2108.08481},
      archivePrefix={arXiv},
      primaryClass={cs.LG},
      doi={https://doi.org/10.5555/3648699.3648788},
      url={https://arxiv.org/abs/2108.08481}, 
}

@article{article,
author = {Adel, Tameem and Valera, Isabel and Ghahramani, Zoubin and Weller, Adrian},
year = {2019},
month = {07},
pages = {2412-2420},
title = {One-Network Adversarial Fairness},
volume = {33},
journal = {Proceedings of the AAAI Conference on Artificial Intelligence},
doi = {10.1609/aaai.v33i01.33012412}
}

@inproceedings{chen2019neural,
author = {Chen, Ricky T. Q. and Rubanova, Yulia and Bettencourt, Jesse and Duvenaud, David},
title = {Neural ordinary differential equations},
year = {2018},
publisher = {Curran Associates Inc.},
booktitle = {Proceedings of the 32nd International Conference on Neural Information Processing Systems},
pages = {6572–6583},
numpages = {12},
location = {Montr\'{e}al, Canada},
series = {NIPS'18}
}

@ARTICLE{2019JCoPh.378..686R,
       author = {{Raissi}, M. and {Perdikaris}, P. and {Karniadakis}, G.~E.},
        title = "{Physics-informed neural networks: A deep learning framework for solving forward and inverse problems involving nonlinear partial differential equations}",
      journal = {Journal of Computational Physics},
         year = {2019},
        month = {Feb},
       volume = {378},
        pages = {686-707},
          doi = {10.1016/j.jcp.2018.10.045},
       adsurl = {https://ui.adsabs.harvard.edu/abs/2019JCoPh.378..686R}
}

@article{Hwang_2022,
   title={Solving PDE-Constrained Control Problems Using Operator Learning},
   volume={36},
   ISSN={2159-5399},
   url={http://dx.doi.org/10.1609/aaai.v36i4.20373},
   DOI={10.1609/aaai.v36i4.20373},
   number={4},
   journal={Proceedings of the AAAI Conference on Artificial Intelligence},
   publisher={Association for the Advancement of Artificial Intelligence (AAAI)},
   author={Hwang, Rakhoon and Lee, Jae Yong and Shin, Jin Young and Hwang, Hyung Ju},
   year={2022},
   month=jun, pages={4504–4512} }

@book{hinze2008optimization,
  title     = {Optimization with PDE Constraints},
  author    = {Hinze, Michael and Pinnau, René and Ulbrich, Michael and Ulbrich, Stefan},
  year      = {2008},
  volume    = {23},
  series    = {Mathematical Modelling: Theory and Applications},
  publisher = {Springer},
  doi       = {10.1007/978-1-4020-8839-1}
}

@book{troeltzsch2010optimal,
  title     = {Optimal Control of Partial Differential Equations: Theory, Methods, and Applications},
  author    = {Tr{\"o}ltzsch, Fredi},
  year      = {2010},
  volume    = {112},
  series    = {Graduate Studies in Mathematics},
  publisher = {American Mathematical Society},
  doi       = {10.1090/gsm/112}
}

@inproceedings{li2020multipole,
  title     = {Multipole Graph Neural Operator for Parametric Partial Differential Equations},
  author    = {Li, Zongyi and Kovachki, Nikola B. and Azizzadenesheli, Kamyar and Liu, Burigede and Bhattacharya, Kaushik and Stuart, Andrew M. and Anandkumar, Anima},
  booktitle = {Advances in Neural Information Processing Systems (NeurIPS)},
  year      = {2020},
  volume    = {33},
  pages     = {6755--6766}
}

@inproceedings{bhattacharya2021model,
  title     = {Model Reduction and Neural Networks for Parametric PDEs},
  author    = {Bhattacharya, Kaushik and Hosseini, Bamdad and Kovachki, Nikola B. and Stuart, Andrew M.},
  booktitle = {Mathematical and Scientific Machine Learning (MSML)},
  year      = {2021},
  volume    = {145},
  series    = {Proceedings of Machine Learning Research},
  pages     = {203--233},
  publisher = {PMLR}
}

@article{zahr2014progressive,
  title={Progressive construction of a parametric reduced-order model for PDE-constrained optimization},
  author={Zahr, Matthew J and Farhat, Charbel},
  year={2014},
  eprint={1407.7618},
journal={International Journal for Numerical Methods in Engineering},
  archivePrefix={arXiv},
  primaryClass={math.OC},
  institution={Institute for Computational and Mathematical Engineering, Stanford University},
  note={https://arxiv.org/abs/1407.7618}
}

@article{dihlmann2015certified,
  title={Certified PDE-constrained parameter optimization using reduced basis surrogate models for evolution problems},
  author={Dihlmann, Markus A and Haasdonk, Bernard},
  journal={Computational Optimization and Applications},
  volume={60},
  pages={753--787},
  year={2015},
  publisher={Springer}
}

@inproceedings{kotary2021end,
  title     = {End-to-End Constrained Optimization Learning: A Survey},
  author    = {Kotary, James and Fioretto, Ferdinando and Van Hentenryck, Pascal and Wilder, Bryan},
  booktitle = {Proceedings of the Thirtieth International Joint Conference on
               Artificial Intelligence, {IJCAI-21}},
  pages     = {4475--4482},
  year      = {2021},
  doi       = {10.24963/ijcai.2021/610},
  url       = {https://doi.org/10.24963/ijcai.2021/610},
}

@inproceedings{park2023self,
  title={Self-supervised primal-dual learning for constrained optimization},
  author={Park, Seonho and Van Hentenryck, Pascal},
  booktitle={Proceedings of the AAAI Conference on Artificial Intelligence},
  volume={37},
  pages={4052--4060},
  year={2023}
}

@article{diamond2016cvxpy,
  title={CVXPY: A Python-embedded modeling language for convex optimization},
  author={Diamond, Steven and Boyd, Stephen},
  journal={The Journal of Machine Learning Research},
  volume={17},
  number={1},
  pages={2909--2913},
  year={2016},
  publisher={JMLR. org}
}

@misc{divito2024learningsolvedifferentialequation,
      title={Learning To Solve Differential Equation Constrained Optimization Problems}, 
      author={Vincenzo Di Vito and Mostafa Mohammadian and Kyri Baker and Ferdinando Fioretto},
      year={2024},
      eprint={2410.01786},
      archivePrefix={arXiv},
      primaryClass={cs.LG},
      url={https://arxiv.org/abs/2410.01786}, 
}

@article{Lu_2021,
   title={Learning nonlinear operators via DeepONet based on the universal approximation theorem of operators},
   volume={3},
   ISSN={2522-5839},
   url={http://dx.doi.org/10.1038/s42256-021-00302-5},
   DOI={10.1038/s42256-021-00302-5},
   number={3},
   journal={Nature Machine Intelligence},
   publisher={Springer Science and Business Media LLC},
   author={Lu, Lu and Jin, Pengzhan and Pang, Guofei and Zhang, Zhongqiang and Karniadakis, George Em},
   year={2021},
   month=mar, pages={218–229} }

@article{MPC,
  title={Review on model predictive control: an engineering perspective},
  author={Max Schwenzer and Muzaffer Ay and Thomas Bergs and Dirk Abel},
  journal={The International Journal of Advanced Manufacturing Technology},
  year={2021},
  volume={117},
  pages={1327 - 1349},
}

@book{Direct-method,
  author = {Betts, John T.},
  title = {Practical Methods for Optimal Control and Estimation Using Nonlinear Programming},
  edition = {2},
  publisher = {SIAM},
  year = {2010}
}

@article{adjoint-method,
author = {Cao, Yang and Li, Shengtai and Petzold, Linda and Serban, Radu},
title = {Adjoint Sensitivity Analysis for Differential-Algebraic Equations: The Adjoint DAE System and Its Numerical Solution},
journal = {SIAM Journal on Scientific Computing},
volume = {24},
number = {3},
pages = {1076-1089},
year = {2003},
}

@article{Crank_Nicolson_1947, title={A practical method for numerical evaluation of solutions of partial differential equations of the heat-conduction type}, volume={43}, DOI={10.1017/S0305004100023197}, number={1}, journal={Mathematical Proceedings of the Cambridge Philosophical Society}, author={Crank, J. and Nicolson, P.}, year={1947}, pages={50–67}}

@article{basis_cite_1, title={Optimal Control Operator Perspective and a Neural Adaptive Spectral Method}, volume={39}, number={14}, journal={Proceedings of the AAAI Conference on Artificial Intelligence}, author={Feng, Mingquan and Chen, Zhijie and Huang, Yixin and Liu, Yizhou and Yan, Junchi}, year={2025}, month={Apr.}, pages={14567-14575} }

@INPROCEEDINGS{basis_cite_2,
  author={Li, Zhexian and Fokas, Athanassios S. and Savla, Ketan},
  booktitle={2024 IEEE 63rd Conference on Decision and Control (CDC)}, 
  title={On linear quadratic regulator for the heat equation with general boundary conditions}, 
  year={2024},
  volume={},
  number={},
  pages={7594-7599},
}

@article{LBFGSB,
author = {Byrd, Richard H. and Lu, Peihuang and Nocedal, Jorge and Zhu, Ciyou},
title = {A Limited Memory Algorithm for Bound Constrained Optimization},
journal = {SIAM Journal on Scientific Computing},
volume = {16},
number = {5},
pages = {1190-1208},
year = {1995},
}

@article{SLSQP,
  title={Algorithm 733: TOMP–Fortran modules for optimal control calculations},
  author={Dieter Kraft},
  journal={ACM Transactions on Mathematical Software (TOMS)},
  year={1994},
  volume={20},
  pages={262 - 281},

}

@book{MLP,
    title={Deep Learning},
    author={Ian Goodfellow and Yoshua Bengio and Aaron Courville},
    publisher={MIT Press},
    year={2016},
}

@article{LSTM,
  title={Long Short-Term Memory},
  author={Sepp Hochreiter and J{\"u}rgen Schmidhuber},
  journal={Neural Computation},
  year={1997},
  volume={9},
  pages={1735-1780},
}
\bibliographystyle{iclr2026_conference}

\newpage

{\Large \textbf{Appendix}}

\section{Classical Methods}
\label{app:classical_methods}
\subsection{Direct Method (Finite-Difference Approach)}
\label{app:dm_baseline}

The direct method treats the simulator as a black box mapping the bounded control vector $\bm u$ to the discrete objective $J(\bm u)$ and employs a standard optimizer to minimize it.
Here the control variable is discretized at each collocation point: $\bm u\in\mathbb{R}^{N_x}$. 

\textbf{Mechanism.}
We use  L-BFGS-B\citep{LBFGSB} with per–coordinate box bounds $u_{\min}\le u_j\le u_{\max}$ to minimize (\ref{eq:disc_obj_no_basis}). No analytic gradients are provided; instead, numerical derivatives are queried via finite differences. With a forward-difference method,
\begin{equation}
\frac{\partial J}{\partial u_j} \;\approx\; 
\frac{J(\bm u+\epsilon_j \bm e_j)-J(\bm u)}{\epsilon_j},
\qquad
\epsilon_j = \eta\,\bigl(1+|u_j|\bigr),
\end{equation}
where $\bm e_j$ is the $j$-th unit vector and $\eta$ is a small scalar (e.g., $\mu = 10^{-6}$). For higher accuracy (to which corresponds a double computational cost), the central-difference method is adopted:
\begin{equation}
\frac{\partial J_h}{\partial u_j} \;\approx\; 
\frac{J_h(\bm u+\epsilon_j \bm e_j)-J_h(\bm u-\epsilon_j \bm e_j)}{2\epsilon_j}.
\end{equation}

\textbf{Computational cost.}
Each gradient evaluation requires $N_x\!+\!1$ simulator calls with forward differences (or $2N_x$ with central differences), where $N_x$ denotes the number of spatial collocation points. Since one simulator call is a complete rollout (\ref{eq:cn_linear_system_no_basis}) over $N_t$ time steps, the complexity per iteration is $\mathcal{O}\big((N_x\!+\!1)\,N_t\,N_x\big)$.
This method is robust and simple, but its cost scales linearly with $N_x$, which becomes prohibitive for dense spatial discretizations.

\subsection{Adjoint Sensitivity Method for Open-Loop Problem}
\label{app:adj_baseline}
The Adjoint-Sensitivity Method efficiently computes exact gradients of PDE-constrained objectives w.r.t.\ high-dimensional control variables. Crucially, the cost of one gradient evaluation is \emph{one forward solve + one backward adjoint solve}, regardless of the number of control parameters. We derive the continuous adjoint for the open-loop, time-invariant Voltage Control Optimization (\ref{exp:voltage}) and then we present the discrete-time adjoint state used in our Crank–Nicolson (CN) implementation.

\textbf{Function spaces \& BCs.}
Consider $u \in L^2(\Omega)$, $y \in L^2([0,T]\!\times\!\Omega)$, and impose homogeneous Neumann BCs on $V$ so that the boundary terms vanish in the variational calculus; the adjoint inherits the same Neumann BCs.

The problem consists of finding $u(x)$ minimizing
\begin{equation}
\label{app:obj_continuous_voltage}
J(u) = \tfrac{1}{2}\!\int_{\Omega}\!\!(y(T, x) - y_{\text{target}}(x))^2\, dx \;+\; \tfrac{\gamma}{2}\!\int_{\Omega}\! u(x)^2\, dx,
\end{equation}
subject to
\begin{subequations} \label{app:pde_continuous_voltage}
\begin{align}
& \frac{\partial y}{\partial t} - D \frac{\partial^2 y}{\partial x^2} + \beta\big(y - y_{\text{ref}}(x)\big) - \alpha u(x) = 0, && \forall (t,x)\in [0,T]\times\Omega, \label{app:pde_voltage_dynamics} \\
& y(0, x) = y_0(x), && \forall x \in \Omega, \label{app:pde_voltage_initial} \\
& \partial_x y(t,x) = 0, && \forall (t,x) \in [0,T]\times\partial\Omega, \label{app:pde_voltage_boundary}
\end{align}
\end{subequations}
where $\gamma>0$ is the regularization weight in the control effort.

\textbf{Lagrangian.} We define
\begin{equation}
\label{app:lagrange_equation_voltage}
\mathcal{L}(u,y,p)=J(u) + \int_{0}^{T}\!\!\int_{\Omega} p(t, x)\!\left[\frac{\partial y}{\partial t} - D \frac{\partial^2 y}{\partial x^2} + \beta\big(y - y_{\text{ref}}(x)\big) - \alpha u(x)\right] dx\,dt,
\end{equation}
where $p (t, x)$ is the adjoint state. By imposing $\nabla_{y} J = 0$ and integrating by parts in $t$ and $x$ (with Neumann BCs) yields the adjoint system.

\textbf{Adjoint PDE.}
\begin{equation}
\label{app:adjoint_pde_voltage}
-\frac{\partial p}{\partial t} - D \frac{\partial^2 p}{\partial x^2} + \beta\, p = 0,
\qquad \partial_x p|_{\partial\Omega}=0.
\end{equation}

\textbf{Terminal condition.}
\begin{equation}
\label{app:adjoint_terminal_voltage}
p(T, x) = y(T, x) - y_{\text{target}}(x).
\end{equation}

\textbf{Gradient w.r.t.\ $u(x)$.}
Because $u$ is time-invariant, variation w.r.t.\ $u$ gives
\begin{equation}
\label{eq:adjoint_gradient}
\nabla_{u(x)} J \;=\; \gamma\,u(x) \;-\; \alpha \int_{0}^{T} p(t, x)\,dt.
\end{equation}
We optimize $\bm u$ with L\,-BFGS\,-B under elementwise bounds $u_{\min}\le u_i\le u_{\max}$; no projection step is needed.


\textbf{Semi-discrete forward dynamics.}
To simplify the notation, let $\bm y(t) = \Big[y(t,x_1) \; \dots \; y(t,x_i) \; \dots \; y(t,x_{N_x})\Big]\in\mathbb{R}^{N_x}$ be the discretized version of $y(t,x)$ on a uniform grid with Neumann BCs,
$\bm{L}\in\mathbb{R}^{N_x\times N_x}$ be the the Neumann Laplacian, and
$\bm u\in\mathbb{R}^{N_x}$ the \emph{static} control vector with elements $u(x_i)$.
Define $\bm{A}:=D\,\bm{L}-\beta\,\bm{I}_{N_x}$ and $\bm s:=\beta\,\bm y_{\text{ref}}$.
Then
\begin{equation}
\label{eq:semi_discrete_no_basis}
\dot{\bm y}(t) \;=\; \bm{A}\,\bm y(t)\;+\;\alpha\,\bm u\;+\;\bm s .
\end{equation}

\textbf{Crank--Nicolson (CN) step.}
With $\Delta t=T/N_t$ and $t_k=k\Delta t$, CN gives
\begin{equation}
\label{eq:cn_linear_system_no_basis}
\big(\bm{I}_{N_x}-\tfrac{\Delta t}{2}\bm{A}\big) \bm y(t_{k+1})
\;=\;
\big(\bm{I}_{N_x}+\tfrac{\Delta t}{2}\bm{A}\big)\, \bm y(t_k)
\;+\;\Delta t\,\alpha\,\bm u
\;+\;\Delta t\,\bm s .
\end{equation}
Equivalently, the affine one-step map is
\begin{align}
\label{eq:cn_map_no_basis}
\bm y(t_{k+1})
&=\bm{A}_{\mathrm{CN}}\,\bm y(t_{k}) \;+\; \bm{B}_{\mathrm{stat}}\,\bm u \;+\; \bm d,\\
\label{B_stat_def}
\bm{A}_{\mathrm{CN}}&=\big(\bm{I}_{N_x}-\tfrac{\Delta t}{2}\bm{A}\big)^{-1}\!\big(\bm{I}_{N_x}+\tfrac{\Delta t}{2}\bm{A}\big),\qquad
\bm{B}_{\mathrm{stat}}=\Delta t\,\big(\bm{I}_{N_x}-\tfrac{\Delta t}{2}\bm{A}\big)^{-1}\,\alpha\,\bm{I}_{N_x},\\
\label{eq:cn_defs_no_basis}
\bm d&=\Delta t\,\big(\bm{I}_{N_x}-\tfrac{\Delta t}{2}\bm{A}\big)^{-1}\,\bm s .
\end{align}

\textbf{Discrete objective.}
With $\bm{M}=\Delta x\,\bm{I}_{N_x}$ we have: 
\begin{equation}
\label{eq:disc_obj_no_basis}
J_h(\bm u)
=\tfrac{1}{2}\,\|\bm y_{N_t}-\bm y_{\text{target}}\|_{\bm{M}}^2
+\tfrac{\gamma}{2}\,\|\bm u\|_{\bm{M}}^2
=\tfrac{1}{2}(\bm y_{N_t}-\bm y_{\text{target}})^\top\bm{M}(\bm y_{N_t}-\bm y_{\text{target}})
+\tfrac{\gamma}{2}\,\bm u^\top\bm{M}\bm u .
\end{equation}

\textbf{Discrete adjoint \& gradient w.r.t.\ static $\bm u$.}
\begin{equation}
\label{eq:disc_adj_no_basis}
\bm\lambda_{N_t}=\bm{M}(\bm y_{N_t}-\bm y_{\text{target}}),\qquad
\bm\lambda_k=\bm{A}_{\mathrm{CN}}^\top\,\bm\lambda_{k+1},\quad k=N_t-1,\dots,0,
\end{equation}
\begin{equation}
\label{eq:grad_u_no_basis}
\nabla_{\bm u} J_h
=\gamma\,\bm{M}\,\bm u
-\sum_{k=0}^{N_t-1}\bm{B}_{\mathrm{stat}}^\top\,\bm\lambda_{k+1}
\;=\;
\gamma\,\bm{M}\,\bm u
-\alpha\,\Delta t\,\big(\bm{I}_{N_x}-\tfrac{\Delta t}{2}\bm{A}\big)^{-\!\top}
\!\left(\sum_{k=0}^{N_t-1}\bm\lambda_{k+1}\right).
\end{equation}

\textbf{Computational cost.}
One forward CN rollout (\ref{eq:cn_linear_system_no_basis}) and one backward adjoint sweep (\ref{eq:disc_adj_no_basis}) yield the exact discrete gradient (\ref{eq:grad_u_no_basis}); then, $\bm u$ is updated with L-BFGS-B under box bounds. In 1D, the CN matrix is time-invariant and can be prefactored once, so that each step reduces to cheap triangular-matrix solver call; the totla cost per iteration is $\mathcal{O}(N_t N_x)$.

\subsection{Time-Varying Adjoint for Closed-Loop Baselines}
\label{app:tv_adj}
The adjoint method extends naturally to a time-varying control $u(t, x)$. The continuous adjoint PDE and terminal/boundary conditions remain as per Eq. (\ref{app:adjoint_pde_voltage})–(\ref{app:adjoint_terminal_voltage}), while the gradients w.r.t. $u(t,x)$ are calculated as:
\begin{equation}
\nabla_{u(t, x)} J \;=\; \gamma\,u(t,x)\;-\;\alpha\,p(t,x).
\end{equation}
We use this time-varying adjoint for the Heat and Burgers experiments (please refer to Section~\ref{sec:Heat} and~\ref{sec:burger}); the discrete forms below match our CN implementations.

\paragraph{Heat optimization problem — CN discretization.}
With $\bm{y}(t_{k+1})=\bm{A}\bm{y}(t_{k})+\bm{B}\bm{u}(t_k)+\bm{g}$ and a quadratic running/terminal costs,
the discrete adjoint equation and per-step gradient are
\begin{align}
\bm\lambda_{N_t} &= \bm{Q}\,(\bm{y}_{N_t}-\bm y_{\text{target}}), \qquad
\bm\lambda_k = \bm{A}^\top \bm\lambda_{k+1} + \bm{T}_Q\,(\bm{y}(t_k)-\bm y_{\text{target}}), \\
\nabla_{\bm{u}(t_k)} J_h &= \bm{R}\,\bm{u}(t_k) + \bm{B}^\top \bm\lambda_{k+1}.
\end{align}
This is the gradient used by LMPC in Section~\ref{sec:Heat}. $\bm{Q}, \bm{T}_Q, \text{ and } \bm{R}$ represent the weight matrices.

\paragraph{Burgers' optimization problem — implicit CN step.}
We write one CN step as the implicit map $\bm y(t_{k+1})=\mathcal{G}(\bm y(t_{k}),\bm{u}(t_k);\Delta t)$ with Dirichlet endpoints enforced at each step. Let
\begin{equation}
A_k \;=\; I-\tfrac{\Delta t}{2}\,J(\bm y(t_{k+1})), \qquad
B_k \;=\; -I-\tfrac{\Delta t}{2}\,J(\bm y(t_{k})),  
\end{equation}
where $J(\bm{y})=-\bm{D}\,\mathrm{diag}(\bm{y})+\nu\,\bm{D}^2$ is the Jacobian of the semi-discrete residual, and $\bm{B}$ stacks the actuator shapes. For running cost
$L_k=\tfrac{\Delta t}{2}\|\bm{y}(t_{k+1})-\bm y_{\text{target}}\|_2^2+\tfrac{\alpha\Delta t}{2}\|\bm{u}(t_k)\|_2^2$,
the closed-loop adjoint/backward sweep is
\begin{align}
A_k^\top \bm{q}_k &= \Delta t\,(\bm{y}(t_{k+1})-\bm y_{\text{target}}), \\
\bm{p}_k &= \Delta t\,(\bm{y}(t_{k})-\bm y_{\text{target}}) - B_k^\top \bm{q}_k, \\
\nabla_{\bm{u}(t_k)} J_h &= \Delta t\big(\alpha\,\bm{u}(t_k) + \bm{B}^\top \bm{q}_k\big),
\end{align}
which are the NMPC equations we implemented and related to the experiments in Section~\ref{sec:burger}. Here $\bm{q}_k$ and $\bm{p}_k$ are adjoint the variables.


\subsection{Model Predictive Control (MPC)}
\label{app:mpc}
In what follows, we introduce the MPC algorithm; to simplify notation we omit the dependency from the spatial variable $x$.
At each time instant $t_k$, MPC solves a finite-horizon problem of $N_p$ steps given inputs
$\{\bm{u}(t_i)\}_{i=k}^{k+N_p-1}$; it applies the first input $\bm{u}(t_k)$, then applies shifts and iterates(receding horizon). 

\paragraph{Linear MPC for the Heat equation.}
With Crank–Nicolson discretization and cosine-basis expansion for the control action, the dynamics are linear:
\begin{equation}
\label{eq:lmpc_dyn}
\bm{y}(t_{i+1}) \;=\; \bm{A}\,\bm{y}(t_{i}) \;+\; \bm{B}\,\bm{c}(t_i) \;+\; \bm{g},
\qquad \bm{y}(t_{i})\in\mathbb{R}^{N_x},\;\; \bm{c}(t_i)\in\mathbb{R}^{M}.
\end{equation}
We solve the convex QP
\begin{equation}
\label{eq:lmpc_qp}
\begin{aligned}
\min_{\{\bm{c}(t_i)\}_{i=k}^{k+N_p-1}} \;&
\sum_{i=k}^{k+N_p-1} \|\bm{y}(t_{i})-\bm y_{\text{target}}\|_{\bm{T}_Q}^2
+ \|\bm{c}(t_i)\|_{\bm{R}}^2
+ \|\bm{y}(t_{k+N_p})-\bm y_{\text{target}}\|_{\bm Q}^2 \\[6pt]
\text{s.t.}\;& (\ref{eq:lmpc_dyn}) \\ 
\;\;& c_{\min}\le \bm{c}(t_i) \le c_{\max}.
\end{aligned}
\end{equation}
Here, $\bm{A}=M_L^{-1}M_R$, $\bm{B}=M_L^{-1}(\Delta t\,\bm{B}_c)$,
$\mathbf{g}=M_L^{-1}(\Delta t\,\mathbf{s})$ (Neumann BCs), $\mathbf{B}_c$ stacks $\cos(j\pi x/X)$; (\ref{eq:lmpc_qp}) is solved  with CVXPY/OSQP \citep{diamond2016cvxpy}. Here $M_L$ and $M_R$ are the CN  matrices of the system, obtained from the 
semi-discrete PDE operator $A_c$:
\begin{equation}
M_L \;=\; I - \tfrac{\Delta t}{2}\,A_c, 
\qquad
M_R \;=\; I + \tfrac{\Delta t}{2}\,A_c.
\end{equation}
This way one CN step writes as:
\begin{equation}
M_L\,\bm y(t_{k+1}) \;=\; M_R\,\bm y(t_k) \;+\; \Delta t\,\bm B_c\,\bm c(t_k) \;+\; \Delta t\,\bm s.
\end{equation}

\paragraph{Nonlinear MPC for the Burgers equation.}
For viscous Burgers, the CN step is nonlinear due to convection term; we write one step as the implicit map
\begin{equation}
\label{eq:nmpc_dyn}
\bm{y}(t_{i+1}) \;=\; \mathcal{G}(\bm{y}(t_{i}),\bm{c}(t_i);\Delta t),
\end{equation}
where $\mathcal{G}$ is the unique CN solution for Burgers with input $\bm{B}\bm{c}(t_i)$
(computed via \texttt{fsolve} on the CN residual and by imposing Dirichlet endpoints at each step).
\begin{equation}
\begin{aligned}
\label{eq:nmpc_loss_simple}
\min_{\{\bm{c}(t_i)\}_{i=k}^{k+N_p-1}} \;&
\sum_{i=k}^{k+N_p-1} \|\bm{y}(t_{i})-\bm y_{\text{target}}\|_{\bm{T}_Q}^2
+ \|\bm{c}(t_i)\|_{\bm{R}}^2
+ \|\bm{y}(t_{k+N_p})-\bm y_{\text{target}}\|_{\bm Q}^2 \\[6pt]
\text{s.t.}\;& \ref{eq:nmpc_dyn}, 
\;\; c_{\min}\le \bm{c}(t_i) \le c_{\max}.
\end{aligned}
\end{equation}

We use single shooting over the full-order model: propagate with (\ref{eq:nmpc_dyn}), optimize with SLSQP\citep{SLSQP} under box bounds, apply $\mathbf{y}(t_k)$, advance one CN step, and iterate. Here $\mathbf{B}$ stacks $\sin((j{+}1)\pi x/X)$ shapes used in the implementation. Notably, ${\mathbf{T}_Q}, {\mathbf{R}}, \text{ and } \mathbf{Q}$ represent the weight matrices, which correspond to the multipliers $\bm \lambda$ in our PDE-OP method.

\paragraph{Choice of controller.}
For the voltage optimization task, we use \emph{Direct} and \emph{Adjoint-method for open loop}. For the heat equation, CN yields a linear PDE, so the finite-horizon problem is a convex QP, which can be solved with \emph{Linear MPC}. For Burgers’ equation, the dynamics are nonlinear and CN produces an implicit nonlinear update, so we use \emph{Nonlinear MPC}, with  NMPC optimizing over an implicit CN step. In both the heat and burgers' tasks, we also use \emph{Adjoint-method for closed loop}, and report its results.







\section{Additional Experimental Results} \label{app:all_experiments}

In this section
we evaluate the runtime and terminal tracking error (MSE) achieved by each method on polynomial target profiles, and report these metrics in Tables \ref{tab:exp_summary_voltage_targets},  \ref{tab:exp_summary_heat_targets}, and \ref{tab:exp_summary_burger_targets}. All methods use the same spatial/temporal discretization and solver settings as described in Appendix \ref{app:hyperparam_settings}.


\begin{figure}[h]
\centering
\includegraphics[width=0.95\linewidth]{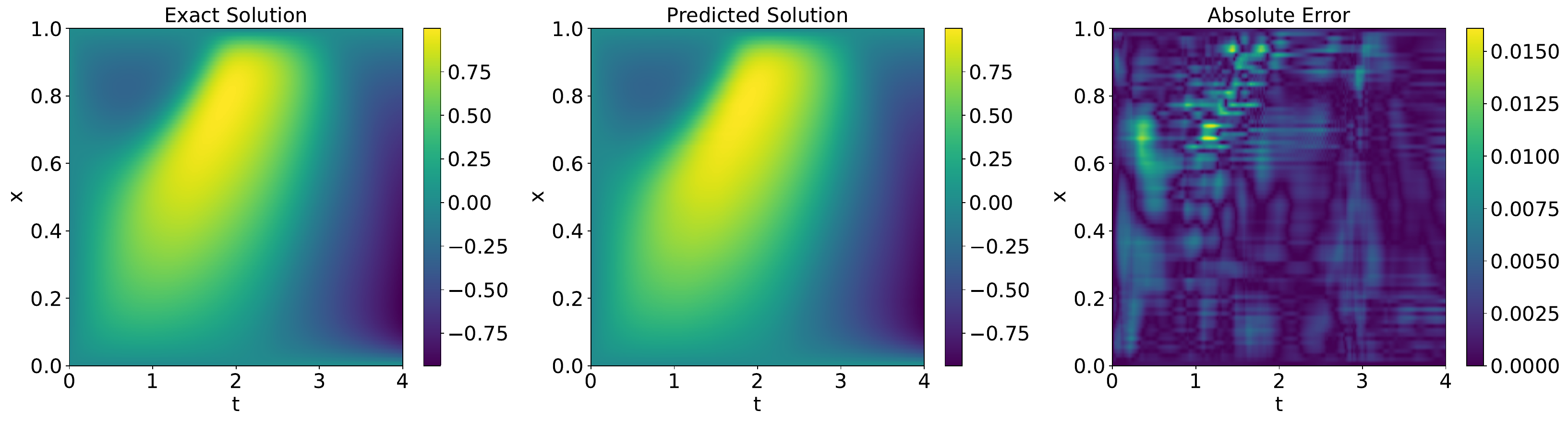}
\caption{Optimal Control of the 1D Burgers' equation with direct modeling; comparison between the solution estimate (center) of PDE-OP's dynamic predictor $\cal Y_\theta$ at test time with the solution obtained with a numerical algorithm (left) and related absolute error (right).}
\label{fig:Exact vs predicted Burger u}
\end{figure}

\subsection{Voltage Control Optimization}
We test each method on additional target profiles $y_{\text{target}}(x)=p_1x+p_2$:
(i) \emph{constant}: $(p_1,p_2)=(0,1)$ so $y_{\text{target}}(x)=1$; 
(ii) \emph{ramp}: $(p_1,p_2)=(1,0.5)$ so $y_{\text{target}}(x)=x+0.5$.
(Equivalently, $y_{\text{target}}(x)=p_1 x + p_2$ with $(p_1,p_2)=(0,1)$ and $(1,0.5)$, respectively.)
Since a ramp target profile is infeasible under zero-flux (Neumann) boundaries, all methods can only produce approximations.

\begin{table}[h]
\centering
\caption{Voltage Optimization Task. For each method and target profile we report runtime (s) and MSE.}
\label{tab:exp_summary_voltage_targets}
\begin{tabular}{@{}l cc cc@{}}
\toprule
\textbf{Methods} & \multicolumn{2}{c}{$y_{\text{target}}(x) = 1$} & \multicolumn{2}{c}{$y_{\text{target}}(x) = x+0.5$} \\ 
\cmidrule(lr){2-3} \cmidrule(lr){4-5}
& \textbf{Runtime} & \textbf{Accuracy} & \textbf{Runtime} & \textbf{Accuracy} \\ \midrule \\[-0.8em]
Direct (Finite Diff.) & 2.422s & 4.6498e-11 & 12.810s & 0.00005 \\
Adjoint-Method & 0.049s  & \textbf{2.8213e-12} & 0.464s  & \textbf{0.00004} \\
\textbf{PDE-OP (ours)}   & \textbf{0.037s}  & 1.5275e-6 & \textbf{0.050s}  & {0.00007} \\ \\[-0.5em] 
\bottomrule
\end{tabular}
\end{table}

\paragraph{Results.}
\emph{Constant target $y_{\text{target}}(x)=1$.} The Adjoint method attains the highest accuracy (MSE $\mathbf{2.82\times 10^{-12}}$) in $0.049$ s, while the Direct method is much slower at $2.422$ s with MSE $4.65\times 10^{-11}$.
\textsc{PDE-OP} is fastest at $\mathbf{0.037}$ s (about $65.5\times$ faster than Direct and $1.32\times$ faster than Adjoint) and reports an MSE of $1.53\times 10^{-6}$.

\emph{Ramp target $y_{\text{target}}(x)=x+0.5$.} The Adjoint method yields the lowest error ($\mathbf{4.0\times 10^{-5}}$), running in $0.464$ s; the Direct method achieves $5.0\times 10^{-5}$ but takes $12.810$ s. \textsc{PDE-OP} runs in $\mathbf{0.050}$ s, about $256\times$ faster than Direct and $9.3\times$ faster than Adjoint method, producing an MSE of $7.0\times 10^{-5}$, which is on same order of magnitude of the baselines.




\subsection{Optimal Control of the 1D Heat Equation}
For the 1D heat equation (closed-loop control), we consider affine targets
\(y_{\text{target}}(x)=p_1 x+p_2\) with \((p_1,p_2)=(0,1)\) and \((1,0.5)\).
Under zero-flux (Neumann) boundary conditions, the ramp profile is not exactly admissible, as its derivative at the boundaries is nonzero; all controllers therefore track its best achievable approximation under these constraint. Here, we compare the proposed PDE-OP model with LMPC and the Adjoint-Sensitivity Method. 

\begin{table}[h]
\centering
\caption{Heat Equation Task. For each method and target profile we report runtime (s) and MSE.}
\label{tab:exp_summary_heat_targets}
\begin{tabular}{@{}l cc cc@{}}
\toprule
\textbf{Methods} & \multicolumn{2}{c}{$y_{\text{target}}(x) = 1$} & \multicolumn{2}{c}{$y_{\text{target}}(x) = x+0.5$} \\ 
\cmidrule(lr){2-3} \cmidrule(lr){4-5}
& \textbf{Runtime} & \textbf{Accuracy} & \textbf{Runtime} & \textbf{Accuracy} \\ \midrule \\[-0.8em]
LMPC & 0.443s & \textbf{2.3289e-14} & 0.562s & \textbf{0.00008} \\
Adjoint-Method & 0.485s  & 6.6612e-11 & 0.552s  & \textbf{0.00008} \\
\textbf{PDE-OP (ours)}   & \textbf{0.223s}  & 1.5210e-6 & \textbf{0.132s}  & {0.00009} \\ \\[-0.5em] 
\bottomrule
\end{tabular}
\end{table}

\paragraph{Results.}
\emph{Constant target $y_{\text{target}}(x)=1$.} LMPC achieves produces accurate terminal profile (MSE $\mathbf{2.33\times 10^{-14}}$) and runs in $0.443$ s; the Adjoint runs in $0.485$ s with MSE $6.66\times 10^{-11}$. \textsc{PDE-OP} is fastest at $\mathbf{0.223}$ s (about $2.0\times$ faster than LMPC and $2.2\times$ than Adjoint) with a still small error of $1.52\times 10^{-6}$.

\emph{Ramp target $y_{\text{target}}(x)=x+0.5$.} LMPC and Adjoint achieves for the highest accuracy (MSE=$\mathbf{8.0\times 10^{-5}}$) with runtimes $0.562$ s and $0.552$ s, respectively. \textsc{PDE-OP} runs only in $\mathbf{0.132}$ s (about $4.3\times$ faster LMPC and $4.2\times$ than the Adjoint), with an MSE of $9.0\times 10^{-5}$ of the same order of magnitude as the baselines.




\subsection{Optimal Control of the  1D Burgers’ Equation}
For the 1D viscous Burgers equation with Dirichlet boundaries $y(t, 0)=y(t,L)=0$, admissible steady targets must satisfy these conditions. We therefore consider (i) a \emph{zero} (constant) target $y_{\text{target}}(x)=0$, and (ii) a \emph{parabolic} target $y_{\text{target}}(x)=p_1\,x(1-x)$ (we use $p_1=2.0$ in experiments). We compare our method with NMPC and time-varying Adjoint baseline. 

\begin{table}[h]
\centering
\caption{Burger Optimization Task. For each method and target profile we report runtime (s) and MSE.}
\label{tab:exp_summary_burger_targets}
\begin{tabular}{@{}l cc cc@{}}
\toprule
\textbf{Methods} & \multicolumn{2}{c}{$y_{\text{target}}(x) = 0$} & \multicolumn{2}{c}{$y_{\text{target}}(x) = 2x(1-x)$} \\ 
\cmidrule(lr){2-3} \cmidrule(lr){4-5}
& \textbf{Runtime} & \textbf{Accuracy} & \textbf{Runtime} & \textbf{Accuracy} \\ \midrule \\[-0.8em]
NMPC & 132.243s & \textbf{5.6760e-14} & 671.033s & \textbf{0.00006} \\
Adjoint-Method & 21.487s  & 5.8290e-11 & 122.116s  & {0.00013} \\
\textbf{PDE-OP (ours)}   & \textbf{0.059s}  & 0.00002 & \textbf{0.073s}  & {0.00031} \\ \\[-0.5em] 
\bottomrule
\end{tabular}
\end{table}

\paragraph{Results.}
\emph{Zero target $y_{\text{target}}(x)=0$.} NMPC reaches nearly machine precision (MSE $\mathbf{5.68\times 10^{-14}}$) but is slow at $132.243$ s; the Adjoint baseline is faster at $21.487$ s with MSE $5.83\times 10^{-11}$. \textsc{PDE-OP} is fastest at $\mathbf{0.059}$ s (about $2.24\times 10^{3}$ vs.\ NMPC and $3.64\times 10^{2}$ vs.\ Adjoint) with a small but larger error of $2.0\times 10^{-5}$.

\emph{Parabolic target $y_{\text{target}}(x)=2x(1\!-\!x)$.} NMPC attains the lowest error (MSE $\mathbf{6.0\times 10^{-5}}$) in $671.033$ s; Adjoint yields $1.3\times 10^{-4}$ at $122.116$ s. \textsc{PDE-OP} runs in $\mathbf{0.073}$ s—about $9.19\times 10^{3}$ faster than NMPC and $1.67\times 10^{3}$ faster than Adjoint—with MSE $3.1\times 10^{-4}$ (same order of magnitude).

\noindent\textit{Takeaway.} For nonlinear Burgers dynamics with Dirichlet BCs, NMPC/Adjoint provide the best MSE, while \textsc{PDE-OP} delivers three to four orders of magnitude lower runtime with modest accuracy loss.




\subsection{Discrete-Time Neural Operator with Direct Control Field \texorpdfstring{$u(x,t)$}{u(x,t)}}
\label{app:dtno_direct_u}

In this section we describe a variant of our discrete-time neural operator that learns \emph{directly} the
spatio–temporal control field \(u(t, x)\) rather than the weights of the basis function as introduced in Section \ref{sec: gen-DON}.
Similarly to the original DeepONet~\citep{Lu_2021}, PDE-OP's model \(\mathcal{Y}_\theta\) factors into a \emph{branch} network that
encodes input functions and a \emph{trunk} network that encodes query coordinates. At each time
\(t_k\), given the current state \(y(t_k, x)\) and the applied control \(u(t_k, x)\), this model
predicts the next state at coordinate \(x\):
\begin{equation}
\label{eq:dtno_update_directu}
\hat{ \bm y}(t_{k+1},x)
\;=\;
\mathcal{Y}_\theta\!\big(t,\,x,\,\hat{\bm y}(t_k, x),\,\hat{\bm {u}}({t_k}, x)\big)
\end{equation}

\textbf{Branch network.}  
The branch net encodes the estimated system state $\hat {\bm y}(t_k, x)$ sampled at $n$ fixed sensor locations
together with estimated control signals $\hat{\bm u}(t_k, x)$, and maps this concatenated input to a $d$-dimensional vector of coefficients
$\bm b\!\left(\hat{\bm y}(t_k, x), \hat{\bm u} (t_k)\right) \in \mathbb{R}^d$.\\    
\textbf{Trunk network.}  
The trunk net encodes the spatial query coordinate $x$, producing a latent feature vector $\bm \gamma(x) \in \mathbb{R}^d$. 
\newline The estimated subsequent (in time) state $\hat {\bm  y}(t_{k+1},x)$ at location $x_i$ is obtained by combining the branch coefficients and trunk features through an inner product:  
\begin{equation}
    \hat { y}(t_{k+1},x_i) \;=\; \sum_{j=1}^d b_j\!\left(\hat {{y}}(t_k, x_i), \hat {\bm u}(t_k, x_i)\right)\,\gamma_j(x_i), \; \; i=1,\dots,n.
\end{equation}

\textbf{Model initialization and training (direct $u$).}
In this setting the dynamic predictor $\mathcal{Y}_\theta$ takes as input the control action
\emph{sampled on the spatial grid}, together with the current state.
$\mathcal{Y}_\theta$ is initialized in a supervised fashion by generating the target trajectories
with a numerical PDE solver under diverse time-varying controls $u(t,x)$.

We sample admissible control sequences $\{u(t_k,x_i)\}$ on the grid
$\{x_i\}_{i=1}^{n}$ from a zero-mean Gaussian random field (GRF) with prescribed spatial
covariance (and optional temporal correlation), after which a clipping operation is applied to meet the actuator bounds
$u_{\min}\le u(t_k,x_i)\le u_{\max}$. Rolling the solver with these inputs produces states
$y(t_k,\cdot)$ satisfying~(\ref{eq:1b})-(\ref{eq:1f}). Each trajectory is converted into
one-step training pairs
\[
\big(y(t_k,x_i),\,u(t_k,x_i)\big)\;\longmapsto\; y(t_{k+1},x_i),
\qquad k=0,\dots,T-1, \quad i=1, \dots, n
\]
Formally, the dataset consists of triples $\mathcal{D}
=\Big\{\big(\bm{y}(t_k, x),\,\bm{u}(t_k, x),\,\bm{y}(t_{k+1}, x)\big)\Big\}$
, with $\bm{y}(t_k, x)=[y(t_k,x_1),\ldots,y(t_k,x_n)]^\top$ and $\bm{u}(t_k, x)=[u(t_k,x_1),\ldots,u(t_k,x_n)]^\top$. $\mathcal{Y}_\theta$ is trained to minimize the following loss
\begin{equation} \label{eq:deeponet_loss_new}
    \min_{\theta} \; \mathbb{E}_{(\bm u, y)\sim\mathcal{D}}
    \Big[ \big\| \hat{y}(t, x) - y(t, x) \big\|^2 \Big],
\end{equation}

\subsection{MPC for Direct Control Field \texorpdfstring{$u(x,t)$}{u(x,t)}}

Here we describe a “direct-$u$” variant for viscous Burgers with Dirichlet boundaries, where the control action is parameterized \emph{at every spatial collocation point}. At each time $t_k$ a vector, the optimization task consists of finding $\bm{u}(t_k, x)\in\mathbb{R}^{N_x}$ (e.g., for a given time instant $t_k$ we have $N_x$ spatial collocation point),  and applying an implicit Crank–Nicolson step to propagate the state in time from $t_k$ to $t_{k+1}$. 
\begin{equation}
\begin{aligned}
\label{eq:nmpc_loss_simple_2}
\min_{\{\bm{u}(t_i)\}_{i=k}^{k+N_p-1}} \;&
\sum_{i=k}^{k+N_p-1} \|\bm{y}(t_{i})-\bm y_{\text{target}}\|_{\bm{T}_Q}^2
+ \|\bm{u}(t_i)\|_{\bm{R}}^2
+ \|\bm{y}(t_{k+N_p})-\bm y_{\text{target}}\|_{\bm Q}^2 \\[6pt]
\text{s.t.}\;& \ref{eq:nmpc_dyn}, 
\;\; u_{\min}\le \bm{u}(t_i) \le u_{\max}.
\end{aligned}
\end{equation}
In case of closed loop (NMPC), a horizon of length $N_p$ yields $N_p\,N_x$ decision variables per iteration; we warm start across receding horizons and use bound-constrained quasi-Newton (e.g., L-BFGS-B/SLSQP).  Here ${\mathbf{T}_Q}, {\mathbf{R}}, \text{ and } \mathbf{Q}$ represent the weight matrices, which correspond to identical with the Lagrange multipliers $\bm \lambda$ of our PDE-OP's training algorithm. 
To compute the gradients, we either backpropagate through the discrete CN integrator using automatic differentiation or use the fully discrete adjoint; both methods are exact but require one forward rollout and one backward sweep over the horizon. This grid-wise formulation removes any basis bias and shows the attainable tracking when the controller has full spatial freedom, but it is
While direct modeling of the control action doesn't require any basis representation, its computational cost  
\emph{higher} w.r.t. to the case in which a basis representation is adopted: the per-step cost scales roughly with $N_p\,N_t\,N_x$, due to line-search/gradient evaluation entailing multiple CN solves. 
We therefore include it as a comprehensive baseline to contextualize the efficiency of our reduced-parameter (mode-based) controllers.

\subsection{Experimental Results For Direct Control Field \texorpdfstring{$u(t,x)$}{u(t,x)}}
\label{app:exp_direct_u}
\begin{figure}[t!]
\centering
\includegraphics[width=.95\linewidth]{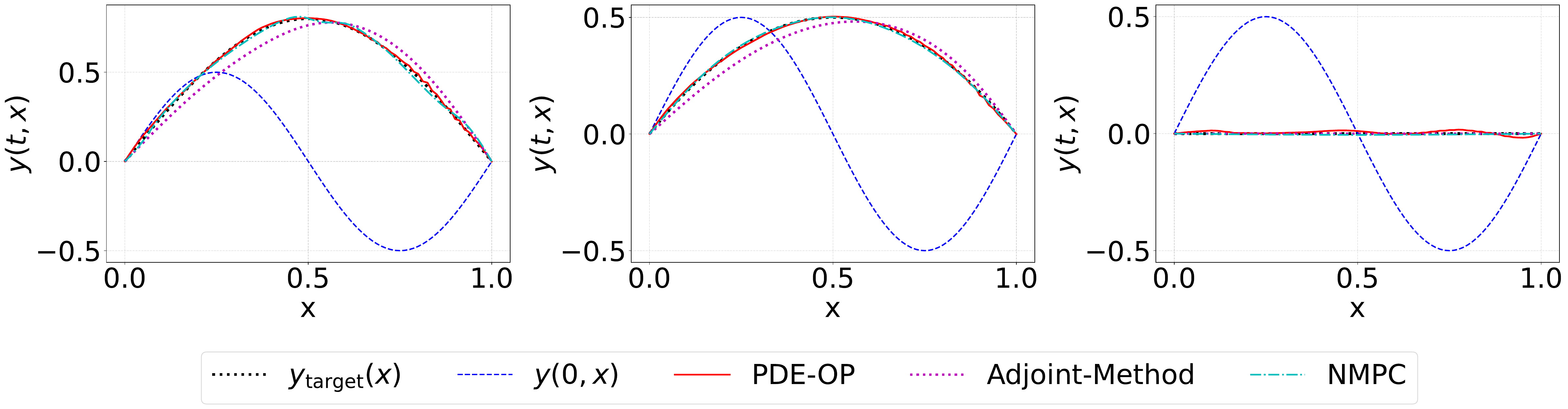}
\caption{Comparison between PDE-OP and each baseline method's solution on three different target profile:  $y_{\text{target}}(x) = 0.8(\sin(x))$ (left),  $y_{\text{target}}(x) = 2x(1-x)$ (center), and $y_{\text{target}}(x) = 0$ (right). We set $N_p = 1$ for NMPC, and $N_p = 10$ for adjoint-method.}
\label{fig:Comparison_burger_direct_u}
\end{figure}

\begin{figure}[t!]
\centering
\includegraphics[width=.95\linewidth]{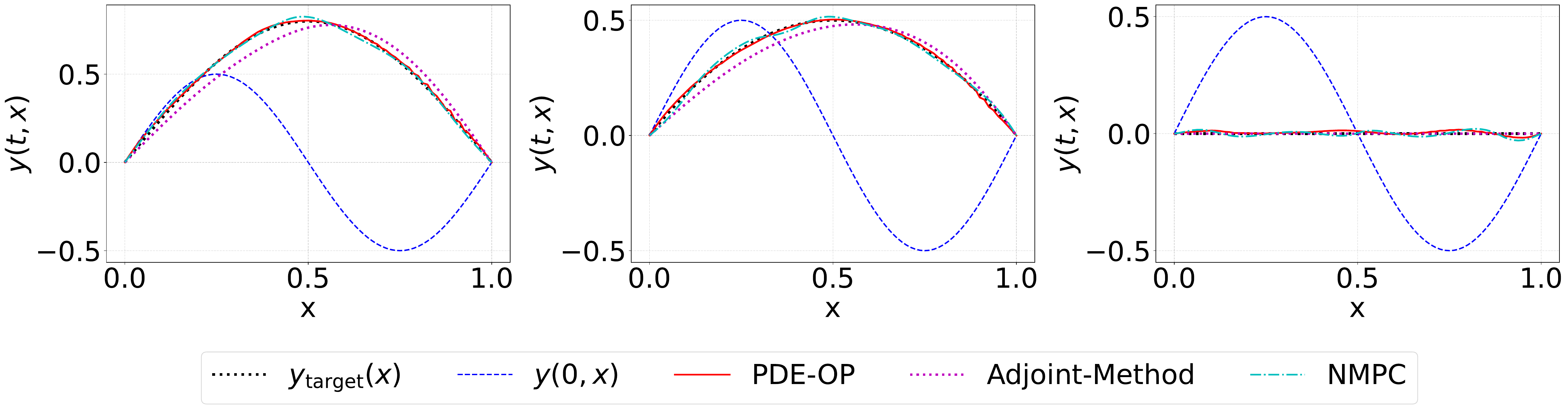}
\caption{Comparison between PDE-OP and each baseline method's solution on three different target profile:  $y_{\text{target}}(x) = 0.8(\sin(x))$ (left),  $y_{\text{target}}(x) = 2x(1-x)$ (center), and $y_{\text{target}}(x) = 0$ (right). We set $N_p = 4$ for NMPC, and $N_p = 20$ for adjoint-method.}
\label{fig:Comparison_burger_direct_u_long}
\end{figure}

When using the direct control field modeling for $u(t,x)$, we evaluate each method on the nonlinear, time-variant optimal control of the 1D Burgers' equation, using the same target profiles presented in Section \ref{sec:burger}. Tables \ref{tab:burgers_dtno_short}-\ref{tab:burgers_dtno_long} report the runtime (s) and terminal tracking error (MSE) of each method. All methods use the same spatial/temporal discretization and solver settings as in \ref{app:hyperparam_settings}. Figure \ref{fig:Exact vs predicted Burger u} shows the predicted voltage (central sub-figure of Fig. \ref{fig:Exact vs predicted Burger u}) of PDE-OP's dynamic component $\mathcal{Y}_\theta$ at test time, which is compared to the solution (left sub-figure) obtained with a numerical PDE-solver. Figures \ref{fig:Comparison_burger_direct_u} and \ref{fig:Comparison_burger_direct_u_long} show the comparison of PDE-OP's predicted terminal state $\hat{y}(T,x)$ and those obtained using the baseline methods, over the same target profiles introduced in the main text. In figure \ref{fig:Comparison_burger_direct_u}, we set $N_p = 1$ for NMPC and $N_p = 10$ for the Adjoint-method, while in figure \ref{fig:Comparison_burger_direct_u_long}, we set $N_p = 4$ for NMPC and $N_p = 20$ for the Adjoint-method.

\begin{table}[h!]
\centering
\caption{ PDE-OP vs.\ classical baselines with short horizon $N_p$. Here we use $N_p{=}1$ for NMPC and $N_p{=}10$ for adjoint. We report runtime (s) and MSE (lower is better).}
\label{tab:burgers_dtno_short}
\begin{tabular}{@{}l cc cc cc@{}}
\toprule
& \multicolumn{2}{c}{$y_{\text{target}}(x) = 2x(1-x)$} & \multicolumn{2}{c}{$y_{\text{target}}(x) = 0.8(\sin(x))$} & \multicolumn{2}{c}{$y_{\text{target}}(x) = 0$}\\
\cmidrule(lr){2-3}\cmidrule(lr){4-5}\cmidrule(lr){6-7}
\textbf{Method} & Runtime & MSE & Runtime & MSE & Runtime & MSE \\ \midrule
NMPC  & 39.254s & \textbf{0.00002} & 76.368s & 0.00014 & 36.483s & 7.9141e$-$06 \\
Adjoint-Method & 47.805s & 0.00124 & 79.644s & 0.00317 & 24.956s & \textbf{2.7618e$-$08} \\ 
\textbf{PDE-OP (ours)} & \textbf{0.267s} & 0.00003 & \textbf{0.256s} & \textbf{0.00008} & \textbf{0.257s} & 0.00008 \\
\bottomrule
\end{tabular}
\end{table}

\paragraph{Results.}
Across all targets, \textsc{PDE-OP} is \emph{two orders of magnitude faster} than classical methods, between $100$–$300\times$ faster then the NMPC and the Adjoint.
In terms of accuracy, for the \emph{sine} target, \textsc{PDE-OP} achieves the lowest MSE; for the \emph{parabola}, NMPC attains the best results with \textsc{PDE-OP} close behind; for the \emph{zero} target, the adjoint method achieves the lowest MSE with \textsc{PDE-OP} producing an MSE of the order of $10^{-5}$. 

\begin{table}[h!]
\centering
\caption{PDE-OP vs.\ classical baselines with longer horizons. Here we use $N_p{=}4$ for NMPC and $N_p{=}20$ for adjoint. We report runtime (s) and MSE (lower is better).}
\label{tab:burgers_dtno_long}
\begin{tabular}{@{}l cc cc cc@{}}
\toprule
& \multicolumn{2}{c}{$y_{\text{target}}(x) = 2x(1-x)$} & \multicolumn{2}{c}{$y_{\text{target}}(x) = 0.8(\sin(x))$} & \multicolumn{2}{c}{$y_{\text{target}}(x) = 0$}\\
\cmidrule(lr){2-3}\cmidrule(lr){4-5}\cmidrule(lr){6-7}
\textbf{Method} & Runtime & MSE & Runtime & MSE & Runtime & MSE \\ \midrule
NMPC      & 826.176s & 0.00016 & 3143.735s & 0.00032 & 470.851s & 0.00061 \\
Adjoint-Method & 186.910s & 0.00129 & 263.891s & 0.00316 & 51.702s & \textbf{4.9213e$-$09} \\
\textbf{PDE-OP (ours)} & \textbf{0.254s} & \textbf{0.00003} & \textbf{0.277s} & \textbf{0.00008} & \textbf{0.956s} & 0.00008 \\
\bottomrule
\end{tabular}
\end{table}

\paragraph{Results.} 
With $N_p{=}4$ (NMPC) and $N_p{=}20$ (adjoint), \textsc{PDE-OP} remains dramatically faster,  three to four orders of magnitude—than classical solvers. 
For the \emph{parabola} target, \textsc{PDE-OP} is $\sim\!3250\times$ faster than NMPC  and $\sim\!740\times$ faster than adjoint, while also reporting the best MSE ($3{\times}10^{-5}$). For the \emph{sine} target, the speedups over classical methods is $\sim\!11300\times$ vs.\ NMPC and $\sim\!950\times$ vs.\ adjoint, while attaining (again) the lowest accuracy ($8{\times}10^{-5}$). 
For the \emph{zero} target, the Adjoint method attains near-perfect regulation ($4.9{\times}10^{-9}$), with \textsc{PDE-OP} is still fastest, running in $0.956\,s$ and approximately $54 \times$ faster than the Adjoint method and $495 \times$ faster than NMPC, while reporting a very small tracking error ($8{\times}10^{-5}$).


\section{Hyperparameter Settings} \label{app:hyperparam_settings}

In Tables \ref{tab:voltage_hparams}, \ref{tab:direct_hparams}, and \ref{tab:adjoint_hparams}, we provide the values of the hyperparameters of PDE-OP model and the baselines baseline methods related to the voltage optimization experiments; Tables \ref{tab:heat_hparams}, \ref{tab:heat_lmpc_hparams}, and \ref{tab:heat_adj_hparams}, report the values of the hyperparameters of PDE-OP model and the baselines baseline methods related to the heat equation experiments; lastly, Tables \ref{tab:burger_hparams}, \ref{tab:burgers_nmpc_hparams}, and \ref{tab:burger_hparams} report the values of the hyperparameters of PDE-OP model and the baselines baseline methods related to the Burgers' equation experiments. For a fair evaluation, each method adopts the same PDE, grid, and horizon parameters $n, N_x, N_t, [u_{\min},u_{\max}], [c_{\min},c_{\max}]$ .

\begin{table}[h!]
\centering
\caption{Voltage equation: simulation parameters and model/training hyperparameters. \\  \textit{MLP} \citep{MLP} a fully connected feedforward network.}
\label{tab:voltage_hparams}
\begin{tabular}{@{}lll@{}}
\toprule
\textbf{Category} & \textbf{Hyperparameter} & \textbf{Value} \\ \midrule
\addlinespace[2pt]
\multirow{8}{*}{PDE \& Simulation} 
  & Domain length $L$                              & $1.0$ \\
  & Final time $T$                  & $5.0$ \\
  & Diffusion $D$                                   & $0.1$ \\
  & Leakage $\beta$                                 & $1.0$ \\
  & Control gain $\alpha$                           & $2.0$ \\
  & Initial state $V(x,0)$                          & $0.0$ \\
  & Reference $V_{\text{ref}}(x)$                   & $1.0$ \\
  & Control bounds $[u_{\min},u_{\max}]$            & $[-1.0,\;1.0]$ \\
\addlinespace[2pt]
\midrule
\multirow{3}{*}{Discretization}
  & Solver grid $(n)$         & $101$ \\
  & Solver steps $(N_t)$         & $101$ \\
  & Number of sensors $(N_x=n)$                & $101$ \\
\addlinespace[2pt]
\midrule
\multirow{11}{*}{PDE-OP's $\cal Y_\theta$}
  & Learning Rate & $5 \times10^{-4}$ \\
  & Number of epochs                        & $1000$ \\
  & Optimizer & Adam \\
  & Batch size                & $1024$ \\
  & Activation function & SiLU \\
  & Lagrange Multiplier $\rho$ & $0.05$ \\
  & Branch net $(b)$: \# layers            & $4$ \\
  & Branch net $(b)$: hidden sizes         & $[512, 512, 512, 512]$ \\
  & Trunk net $(\gamma)$: \# layers             & $4$ \\
  & Trunk net $(\gamma)$: hidden sizes          & $[512, 512, 512, 512]$  \\
  & Feature dimension $(d)$ & $512$ \\  
  
\addlinespace[2pt]
\midrule
\multirow{7}{*}{PDE-OP's $\mathcal{U}_\omega$}
  & Learning rate & $1 \times10^{-3}$ \\
  & Number of epochs            & $1000$ \\
  & Optimizer & AdamW \\
  & Batch size   & $256$ \\
  & Activation function & RELU \\
  & MLP: \# layers  & $3$ \\
  & MLP: hidden sizes & $[256, 256, 256]$
 \\
\bottomrule
\end{tabular}
\end{table}

\begin{table}[h!]
\centering
\caption{Voltage equation: Direct(finite-difference) method parameters.}
\label{tab:direct_hparams}
\begin{tabular}{@{}lll@{}}
\toprule
\textbf{Category} & \textbf{Hyperparameter} & \textbf{Value / Note} \\ \midrule
Optimizer & Algorithm & L-BFGS-B (box bounds $[u_{\min},u_{\max}]$) \\
& Max iterations & 100 \\
& Tolerances & $\texttt{ftol}=10^{-6}$ (and/or $\texttt{gtol}$) \\
& Initialization & $\bm u(t_0)=\mathbf{0}$ (uniform) \\
Finite diff. & Stencil & forward (optionally central) \\
& Step size & $\epsilon_j=\eta(1+|u_j|)$, $\eta=10^{-6}$ \\
Cost eval & Rollout solver & CN\\
& Runtime note & $N_x{+}1$ evals/grad (or $2N_x$ for central) \\ 
\bottomrule
\end{tabular}
\end{table}

\begin{table}[h!]
\centering
\caption{Voltage equation: Adjoint-method parameters.}
\label{tab:adjoint_hparams}
\begin{tabular}{@{}lll@{}}
\toprule
\textbf{Category} & \textbf{Hyperparameter} & \textbf{Value / Note} \\ \midrule
Optimizer & Algorithm & L-BFGS-B (box bounds $[u_{\min},u_{\max}]$) \\
& Max iterations & 100 \\
& Tolerances & $\texttt{ftol}=10^{-6}$ (and/or $\texttt{gtol}$) \\
& Initialization & $\bm u(t_0)=\mathbf{0}$ \\
Adjoint solves & Forward stepper & CN \\
& Backward sweep & discrete adjoint (one pass) \\
Numerics & Linear solves & direct (LU) or iterative tol $10^{-10}$ \\
\bottomrule
\end{tabular}
\end{table}

\begin{table}[h]
\centering
\caption{Heat equation: simulation parameters and model/training hyperparameters. \\ \textit{LSTM} \citep{LSTM} denotes a Long Short-Term Memory network.}
\label{tab:heat_hparams}
\begin{tabular}{@{}lll@{}}
\toprule
\textbf{Category} & \textbf{Hyperparameter} & \textbf{Value} \\ \midrule
\addlinespace[2pt]
\multirow{7}{*}{PDE \& Simulation} 
  & Domain length $L$                              & $1.0$ \\
  & Final time $T$ & $1.0$ \\
  & Diffusion $D$                                   & $0.1$ \\
  & Leakage $\beta$                                 & $0.5$ \\
  & Control gain $\alpha$                           & $2.0$ \\
  & Initial state $V(x,0)$                          & $0.0$ \\
  & Reference $V_{\text{ref}}(x)$                   & $0.0$ \\
  & Weight bounds $[c_{\min},c_{\max}]$            & $[-1.0,\;1.0]$\\
\addlinespace[2pt]
\midrule
\multirow{4}{*}{Discretization}
  & Solver grid $(n)$         & $41$ \\
  & Solver steps $(N_t)$         & $41$ \\
  & Number of sensors $(N_x=n)$                & $41$ \\
  & Number of basis functions $(M)$ & $6$ \\
\addlinespace[2pt]
\midrule
\multirow{10}{*}{PDE-OP's $\cal Y_\theta$}
  & Learning Rate & $1 \times10^{-3}$ \\
  & Number of epochs                        & $2000$ \\
  & Optimizer & AdamW \\
  & Batch size                & $128$ \\
  & Activation function & RELU \\
  & Branch net $(b)$: \# layers            & $4$ \\
  & Branch net $(b)$: hidden sizes         & $[512, 512, 512, 512]$ \\
  & Trunk net $(\gamma)$: \# layers             & $4$ \\
  & Trunk net $(\gamma)$: hidden sizes          & $[512, 512, 512, 512]$  \\
  & Feature dimension $(d)$ & $512$ \\ 
\addlinespace[2pt]
\midrule
\multirow{7}{*}{PDE-OP's $\mathcal{U}_\omega$}
  & Learning rate & $1 \times10^{-4}$ \\
  & Number of epochs            & $5000$ \\
  & Optimizer & AdamW \\
  & Batch size   & $2048$ \\
  & Activation function & RELU \\
  & LSTM: \# layers  & $2$ \\
  & LSTM: hidden sizes & $[256, 256]$ 
 \\
\bottomrule
\end{tabular}
\end{table}

\begin{table}[h]
\centering
\caption{Heat equation: Linear MPC (LMPC) parameters. $(^*)$ The basis function are used by each method.}
\label{tab:heat_lmpc_hparams}
\begin{tabular}{@{}lll@{}}
\toprule
\textbf{Category} & \textbf{Parameter} & \textbf{Value / Note} \\ \midrule
Horizon     & Prediction length $N_p$            & $10$ \\
Dynamics    & State update                        & $\bm{y}(t_{k+1})=\bm{A}\bm{y}(t_{k})+\bm{B}\bm{u}(t_k)+\bm{g}$ \\
            & Discretization                      & Crank--Nicolson (linear) \\
Actuation   & Basis$^*$                              & $\cos(j\pi x/L)$, $j=0,\dots,m-1$ \\
Solver      & QP solver                           & OSQP (CVXPY), warm start \\
Stopping    & Tolerances                          & default OSQP (primal/dual res.) \\
\bottomrule
\end{tabular}
\end{table}

\begin{table}[h]
\centering
\caption{Heat equation: Adjoint method parameters.}
\label{tab:heat_adj_hparams}
\begin{tabular}{@{}lll@{}}
\toprule
\textbf{Category} & \textbf{Parameter} & \textbf{Value / Note} \\ \midrule
Horizon     & Prediction length $N_p$            & $10$ \\
Forward     & Integrator                      & Crank--Nicolson (linear), same $\bm{A},\bm{B},\bm{g}$ \\
Adjoint     & Backward recursion              & $\bm\lambda_{N_t} = \bm{Q}\,(\bm{y}_{N_t}-\bm y_{\text{target}})$ \\
&               &
$\ \bm\lambda_k = \bm{A}^\top \bm\lambda_{k+1} + \bm{T}_Q\,(\bm{y}(t_k)-\bm y_{\text{target}})$ \\
Gradient    & Per step                        & $\nabla_{\bm{u}(t_k)} J_h = \bm{R}\,\bm{u}(t_k) + \bm{B}^\top \bm\lambda_{k+1}.$ \\
Optimizer   & Algorithm                        & L-BFGS-B, warm start \\
Stopping    & Max iters / tolerances          & 100; $\texttt{ftol}=10^{-6}$ \\
\bottomrule
\end{tabular}
\end{table}

\begin{table}[h]
\centering
\caption{Burgers' equation: simulation parameters and model/training hyperparameters. \\ \textit{LSTM}\citep{LSTM} denotes a Long Short-Term Memory network.}
\label{tab:burger_hparams}
\begin{tabular}{@{}lll@{}}
\toprule
\textbf{Category} & \textbf{Hyperparameter} & \textbf{Value} \\ \midrule
\addlinespace[2pt]
\multirow{2}{*}{PDE \& Simulation} 
  & Domain length $L$                              & $1.0$ \\
  & Final time $T$ & $4.0$ \\
  & Viscosity $\nu$                                   & $0.03$ \\
  & Weight bounds $[c_{\min},c_{\max}]$            & $[-1.0,\;1.0]$\\
\addlinespace[2pt]
\midrule
\multirow{4}{*}{Discretization}
  & Solver grid $(n)$         & $81$ \\
  & Solver steps $(N_t)$         & $201$ \\
  & Number of sensors $(N_x=n)$                & $81$ \\
  & Number of basis functions $(M)$ & $4$ \\
\addlinespace[2pt]
\midrule
\multirow{10}{*}{PDE-OP's $\cal Y_\theta$}
  & Learning Rate & $1 \times10^{-3}$ \\
  & Number of epochs                        & $2000$ \\
  & Optimizer & AdamW \\
  & Batch size                & $512$ \\
  & Activation function & RELU \\
  & Branch net $(b)$: \# layers            & $3$ \\
  & Branch net $(b)$: hidden sizes         & $[256, 256, 256]$ \\
  & Trunk net $(\gamma)$: \# layers             & $5$ \\
  & Trunk net $(\gamma)$: hidden sizes          & $[256, 256, 256, 256, 256]$  \\
  & Feature dimension $(d)$ & $128$ \\ 
\addlinespace[2pt]
\midrule
\multirow{7}{*}{PDE-OP's $\mathcal{U}_\omega$}
  & Learning rate & $1 \times10^{-4}$ \\
  & Number of epochs            & $5000$ \\
  & Optimizer & AdamW \\
  & Batch size   & $2048$ \\
  & Activation function & RELU \\
  & LSTM: \# layers  & $2$ \\
  & LSTM: hidden sizes & $[256, 256]$ 
 \\
\bottomrule
\end{tabular}
\end{table}

\begin{table}[h]
\centering
\caption{Burgers’ equation: Nonlinear MPC (NMPC) parameters. $(^*)$ The basis function are used by each method.}
\label{tab:burgers_nmpc_hparams}
\begin{tabular}{@{}lll@{}}
\toprule
\textbf{Category} & \textbf{Parameter} & \textbf{Value / Note} \\ \midrule
Horizon     & Prediction length $N_p$          & e.g.\ $10$ (receding horizon) \\
Dynamics    & Implicit step (CN)               & $\bm {y}(t_{k+1}) \;=\; \mathcal{G}(\bm{y}(t_{k}),\bm{c}(t_k))$ \\
Actuation   & Basis$^{*}$                           & $\sin\!\big((j{+}1)\pi x/X\big)$, $j=0,\dots,M-1$ \\
Solver      & Nonlinear optimizer               & SLSQP (single-shooting), box bounds; warm start by shift \\
CN inner    & Newton solve per step             & tolerance $10^{-10}$, max iters $25$ \\
\bottomrule
\end{tabular}
\end{table}

\begin{table}[h]
\centering
\caption{Burgers’ equation: Adjoint method parameters.}
\label{tab:burgers_adj_hparams}
\begin{tabular}{@{}lll@{}}
\toprule
\textbf{Category} & \textbf{Parameter} & \textbf{Value / Note} \\ \midrule
Horizon     & Prediction lentgh           $N_p$  & 10 \\
Forward     & Implicit CN dynamics            & as in NMPC, Dirichlet enforced each step \\
Jacobian    & Semi-discrete residual          & $J(\bm{y})=-\bm{D}\,\mathrm{diag}(\bm{y})+\nu\,\bm{D}^2$ \\
Adjoint     & Backward sweep                   & $A_k^\top \bm{q}_k = \Delta t\,(\bm{y}(t_{k+1})-\bm y_{\text{target}})$; \\
            &                                  & $\bm{p}_k = \Delta t\,(\bm{y}(t_{k})-\bm y_{\text{target}}) - B_k^\top \bm{q}_k$ \\
Gradient    & Per-step control gradient        & $\nabla_{\bm{u}(t_k)} J= \Delta t\big(\alpha\,\bm{u}(t_k) + \bm{B}^\top \bm{q}_k\big),$ \\
Optimizer   & Outer optimizer                  & L-BFGS-B, warm start \\

\bottomrule
\end{tabular}
\end{table}

\end{document}